%% file: main-for-arXiv.tex
\documentclass[10pt,twocolumn,letterpaper]{article}

\usepackage{cvpr}              

\input{preamble}

\definecolor{cvprblue}{rgb}{0.21,0.49,0.74}

\def\paperID{981} 
\def\confName{CVPR}
\def\confYear{2024}

\title{\method: Instructing Video Diffusion Models with Human Feedback}

\author{%
    Hangjie Yuan\textsuperscript{\rm 1}\thanks{
    Work conducted during his research internships at DAMO Academy.} \hspace{.1in} 
    Shiwei Zhang\textsuperscript{\rm 2} \hspace{.1in} 
    Xiang Wang\textsuperscript{\rm 2} \hspace{.1in} 
    Yujie Wei\textsuperscript{\rm 2} \hspace{.1in} 
    Tao Feng\textsuperscript{\rm 3} \\
    Yining Pan\textsuperscript{\rm 4} \hspace{.1in} 
    Yingya Zhang\textsuperscript{\rm 2} \hspace{.1in} 
    Ziwei Liu\textsuperscript{\rm 5} \hspace{.1in} 
    Samuel Albanie\textsuperscript{\rm 6} \hspace{.1in} 
    Dong Ni\textsuperscript{\rm 1}\thanks{Corresponding author.}  \vspace{.2cm} \\
    \textsuperscript{\rm 1}Zhejiang University \hspace{.2in} 
    \textsuperscript{\rm 2}Alibaba Group \hspace{.2in} 
    \textsuperscript{\rm 3}Tsinghua University \\
    \textsuperscript{\rm 4}Singapore University of Technology and Design \hspace{.2in}
    \textsuperscript{\rm 5}S-Lab, Nanyang Technological University \hspace{.2in} \\
    \textsuperscript{\rm 6}CAML Lab, University of Cambridge
    \vspace{.2cm} \\
    \texttt{Project page: \url{https://instructvideo.github.io/}}
}

\begin{document}
\maketitle
\input{sec/00-abs}
\input{sec/01-intro}

\input{sec/02-related-work}
\input{sec/03-method}

\input{sec/04-exp}

\input{sec/05-conclusion}
{
    \small
    \bibliographystyle{ieeenat_fullname}
    \bibliography{main}
}
\input{sec/06-supp}

\end{document}

%% file: preamble.tex
\usepackage{bm}
\usepackage{bbm}
\usepackage{amsthm,amsmath,amssymb}
\usepackage{mathrsfs}
\usepackage{booktabs}
\usepackage{multirow}
\usepackage{bbding}
\usepackage{makecell}
\usepackage{enumitem}
\usepackage{caption}
\usepackage{subcaption}
\usepackage{graphicx}
\usepackage[detect-none]{siunitx}
\usepackage{wrapfig}
\usepackage{adjustbox}
\usepackage{color}
\usepackage{colortbl}
\usepackage[accsupp]{axessibility}
\usepackage{pifont}
\usepackage{makecell}

\definecolor{citecolor}{RGB}{75, 166, 154}
\usepackage[pagebackref=true,breaklinks=true,letterpaper=true,colorlinks,citecolor=citecolor,bookmarks=false]{hyperref}
\usepackage[capitalize]{cleveref}
\crefname{section}{Sec.}{Secs.}
\Crefname{section}{Section}{Sections}
\Crefname{table}{Table}{Tables}
\crefname{table}{Tab.}{Tabs.}

\usepackage{colortbl} 

\usepackage[dvipsnames]{xcolor}

\newlength\savewidth\newcommand\shline{\noalign{\global\savewidth\arrayrulewidth
  \global\arrayrulewidth 1pt}\hline\noalign{\global\arrayrulewidth\savewidth}}

\newcommand{\method}{\texttt{InstructVideo}\xspace}


%% file: sec/00-abs.tex
\begin{abstract}

Diffusion models have emerged as the de facto paradigm for video generation. 
However, their reliance on web-scale data of varied quality often yields results that are visually unappealing and misaligned with the textual prompts. 
To tackle this problem, we propose \method to instruct text-to-video diffusion models with human feedback by reward fine-tuning. 
\method has two key ingredients: 
\textbf{1)} To ameliorate the cost of reward fine-tuning induced by generating through the full DDIM sampling chain, we recast reward fine-tuning as editing. 
By leveraging the diffusion process to corrupt a sampled video, \method requires only partial inference of the DDIM sampling chain, reducing fine-tuning cost while improving fine-tuning efficiency.
\textbf{2)} To mitigate the absence of a dedicated video reward model for human preferences, we repurpose established image reward models, \textit{e.g.}, HPSv2. 
To this end, we propose Segmental Video Reward, a mechanism to provide reward signals based on segmental sparse sampling, and Temporally Attenuated Reward, a method that mitigates temporal modeling degradation during fine-tuning. 
Extensive experiments, both qualitative and quantitative, validate the practicality and efficacy of using image reward models in \method, significantly enhancing the visual quality of generated videos without compromising generalization capabilities.
Code and models will be made publicly available.

\end{abstract}

%% file: sec/01-intro.tex
\vspace{-.4cm}
\section{Introduction} \label{sec:intro}

The emergence of diffusion models~\cite{sohl2015Diffusion_model,zhang2022gddim,ho2020denoising_DDPM,song2020score_generative_SDE} has significantly boosted generation quality across a wide range of media content~\cite{wang2023modelscopeT2V,kong2021diffwave,ho2022video_diffusion_models,rombach2022stable_diffusion}.
This generation paradigm has shown promise for video generation~\cite{2023videocomposer,wang2023modelscopeT2V,blattmann2023align,ho2022video_diffusion_models,ho2022imagenvideo}, despite the challenges of working with high-dimensional data.
While diffusion models are one factor driving progress, the scaling of training datasets has also played a key role~\cite{singer2022make-a-video,wang2023lavie}.
However, despite recent progress, the visual quality of generated videos still leaves room for improvement~\cite{wang2023modelscopeT2V,wu2023tune-a-video}.
A significant contributing factor to this issue is the varying quality of web-scale data employed during pre-training~\cite{Bain21Webvid,schuhmann2021laion}, which can yield models capable of generating content that is visually unappealing, toxic and misaligned with the prompt.

\begin{figure}[t]
\centering
    \includegraphics[width=.48\textwidth]{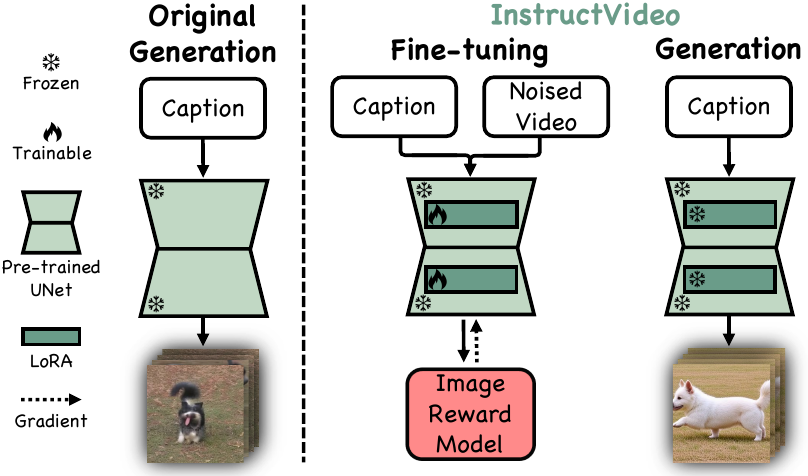}
    \vspace{-0.6cm}
    \caption{\small 
        \textbf{Overview of the \method framework}.
        Our method performs efficient fine-tuning on sampled video-text pairs, instructed by human preferences in image reward models.
    }
\vspace{-0.4cm}
\label{fig:teaser}
\end{figure}


While \textit{aligning model outputs with human preferences} has proven highly effective for control~\cite{christiano2017deep}, text generation~\cite{leike2018scalable_agent_align,askell2021general_language_alignment,ouyang2022InstructGPT,openai2023gpt4,touvron2023llama} and image generation~\cite{lee2023AlignT2I,xu2023ImageReward,wu2023HPSv1}, it remains a notion unexplored in video diffusion models.
The most widely-adopted methods for aligning models with human preferences include off-line reinforcement learning (RL)~\cite{levine2020offlineRL,ouyang2022InstructGPT,black2023DDPO} and direct reward back-propagation~\cite{clark2023DRaFT,prabhudesai2023AlignProp}.
Typically, this entails training a reward model on manually annotated datasets that is then subsequently used to fine-tune the pre-trained generative model.

Two major challenges arise when seeking to align video generation models with human preferences:
\textbf{1)} The optimization process for optimizing human preferences is computationally demanding, often requiring video generation from textual inputs. 
While video generation pre-training using DDPM~\cite{ho2020denoising_DDPM} requires only a single-step inference for every iteration, reward optimization requires a 50-step DDIM inference~\cite{song2020DDIM}.
\textbf{2)} The curation of a large annotated dataset to capture human preferences of videos is labor-intensive, while the computation- and memory-intensive demands of utilizing ViT-H~\cite{wu2023HPSv2} or ViT-L~\cite{xu2023ImageReward}-based computational alternatives to evaluate the entire video are high.

To surmount these mentioned challenges, we propose \method, a model that efficiently instructs text-to-video diffusion models to follow human feedback, as illustrated in~\cref{fig:teaser}.
Regarding the first challenge of the demanding reward fine-tuning process caused by generating through the full DDIM sampling chain, we \textit{recast the problem of reward fine-tuning as an editing procedure}.
This reformulation requires only partial inference of the DDIM sampling chain, thereby reducing computational demands while improving fine-tuning efficiency.
Drawing inspiration from established editing workflows in diffusion models~\cite{meng2021SDedit,couairon2022Diffedit,ceylan2023Pix2video,chai2023Stablevideo,qi2023fatezero,zhao2023controlvideo,liu2023video-p2p,geyer2023tokenflow,molad2023dreamix}, where primary visual content is initially corrupted with noise and then reshaped by a target prompt, our method focuses on refining coarse and structural videos into more detailed and nuanced outputs. 
This contrasts with previous methods~\cite{lee2023AlignT2I,black2023DDPO,clark2023DRaFT} that generate results directly from text. 
Such blurry and structural videos, serving as the starting point for reward fine-tuning, are procured by a simple diffusion process with negligible cost.
During generation, the optimized model retains the capability to produce videos directly from textual inputs.
In conjunction with back-propagation truncation of the sampling chain, we make reward fine-tuning on text-to-video diffusion models computationally attainable and effective.

Regarding the second challenge (the lack of a reward model tailored for video generation), we postulate that the visual excellence of a video is tied to both the quality of its individual frames and the fluidity of motion across consecutive frames.
To this end, we resort to off-the-shelf image reward models, \textit{e.g.}, HPSv2~\cite{wu2023HPSv2}, to ascertain frame quality.
Drawing inspiration from temporal segment networks~\cite{wang2016TSN}, we propose Segmental Video Reward (SegVR), which strategically evaluates video quality based on a subset of sparely sampled frames.
By providing sparse reward signals, SegVR offers dual benefits: it not only ameliorates computational burden but also mitigates temporal modeling collapse.
On the other hand, although LoRA~\cite{hu2021lora} is adopted by default to retain the capability to generate temporally smooth videos, SegVR still leads to videos with visual artifacts, such as structure twitching and color jittering.
To mitigate this, we propose Temporally Attenuated Reward (TAR), which operates under the hypothesis that central frames should be assigned paramount importance, with emphasis tapering off towards peripheral frames.
This strategic allocation of importance across frames ensures a more stable and visually coherent video generation process.

As part of our pioneering effort to align video diffusion models with human preferences, we conduct extensive experiments to assess the practicality and efficacy of integrating image reward models within \method. 
Our findings reveal that \method markedly enhances the visual quality of generated videos without sacrificing the model's generalization capabilities, setting a new precedent for future research in video generation.

%% file: sec/02-related-work.tex
\section{Related Work} \label{sec:related}

\begin{figure*}[t]
\centering
    \includegraphics[width=1.\textwidth]{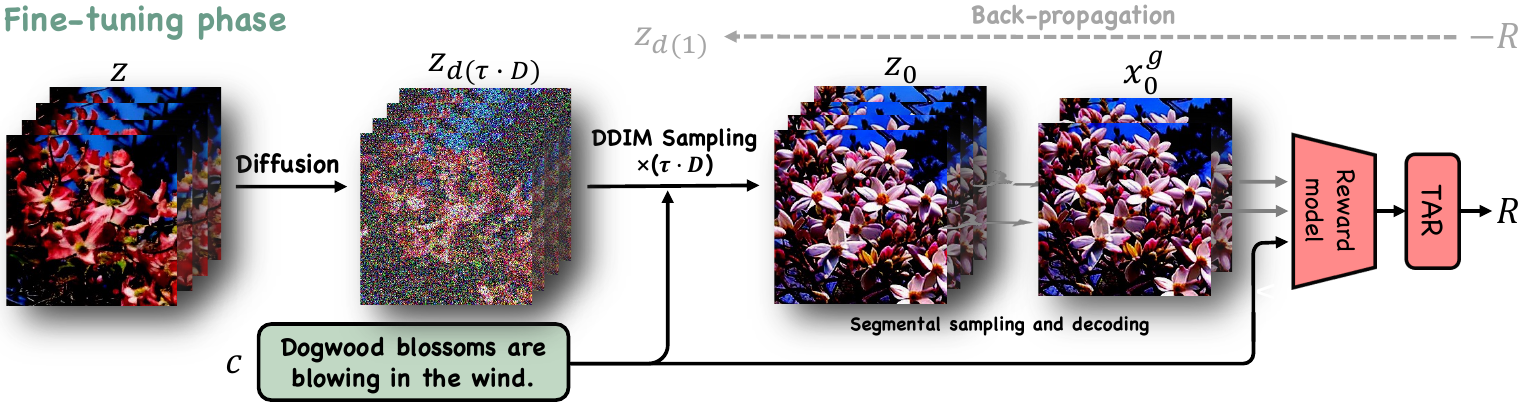}
    \vspace{-0.6cm}
    \caption{\small 
        \textbf{The reward fine-tuning framework of \method}.
        During fine-tuning, we sample video-text pairs and apply a diffusion process to corrupt the videos to a noise level $\tau$.
        Subsequently, we perform partial inference of the DDIM sampling chain to obtain the human preference edited videos.
        By utilizing SegVR and TAR, we can leverage image reward models to perform reward fine-tuning for video generation.
        The VQGAN encoder and decoder are omitted for clarity.
        In this example, the blurry video $\bm{z}$ is edited guided by human preferences, producing a result that highlights the vibrancy and structure of the dogwood blossoms.
    }
\vspace{-0.3cm}
\label{fig:pipeline}
\end{figure*}

\noindent\textbf{Video generation via diffusion models.}
Early efforts at video generation focused on GANs~\cite{pan2017TGANs-C,li2018video_GAN_VAE,villegas2022phenaki_GAN,skorokhodov2022stylegan-v,hong2022cogvideo,yu2022video_implicit_GAN,wu2022nuwa,fu2023tell_me} and VAEs~\cite{li2018video_GAN_VAE,mittal2017sync-draw,yan2021videogpt}. 
However, due to the complexity of jointly modeling spatio-temporal dynamics, generating videos from texts remains an unresolved challenge.
Recent methods for video generation aim to mitigate this by utilizing the de facto generation method, \textit{i.e.}, diffusion models~\cite{sohl2015Diffusion_model,ho2020denoising_DDPM,zhang2022gddim,song2020DDIM}, for generating videos with diversity and fidelity~\cite{ho2022imagenvideo,ho2022video_diffusion_models,zhang2023show-1,wang2023lavie,xing2023SimDA,liu2023dual_T2V,gu2023reuse_diffuse,blattmann2023align,yu2023PVDM,zhou2022magicvideo,an2023latent-shift,wang2023videofactory,esser2023gen-1,ma2023follow_your_pose,yin2023dragnuwa,chen2023MCDiff,xing2023make-your-video,he2023animate-a-story,ruan2023mm-diffusion,tang2023any-to-any,zhu2023moviefactory,chen2023videocrafter1,zhang2023i2vgen-xl,ge2023preserve_correlation,khachatryan2023text2video-zero,chen2023videodreamer,zhao2023motiondirector,ma2023optimal-noise-pursuit,xing2023dynamicrafter,he2022lvdm,chen2023seine,qing2023HiGen,wei2023dreamvideo,wang2023videolcm} and scaling up the pre-training data or model architecture~\cite{singer2022make-a-video,hong2022cogvideo,wu2021godiva,wang2023lavie,ho2022imagenvideo}.
VDM~\cite{ho2022video_diffusion_models} represents a pioneering work that extended image diffusion models to video generation.
Owing to the computation-intensive nature of diffusion models, follow-up research sought to reduce overhead by leveraging the latent space~\cite{rombach2022stable_diffusion}, \textit{e.g.}, ModelScopeT2V~\cite{wang2023modelscopeT2V}, Video LDM~\cite{blattmann2023align}, MagicVideo~\cite{zhou2022magicvideo} and SimDA~\cite{xing2023SimDA}, \textit{etc.}.
To enable more controllable generation, further efforts introduce spatio-temporal conditions~\cite{chen2023controlavideo,2023videocomposer,yin2023dragnuwa,esser2023gen-1,lapid2023gd-vdm,xing2023make-your-video}, \textit{e.g.}, VideoComposer~\cite{2023videocomposer}, Gen-1~\cite{esser2023gen-1}, DragNUWA~\cite{yin2023dragnuwa}, \textit{etc.}.
However, generating videos that adhere to human preferences remains a challenge. 

\noindent\textbf{Human preference model.}
Understanding human preference in visual content generation remains challenging~\cite{kirstain2023Pick-a-pic,xu2023ImageReward,wu2023HPSv1,wu2023HPSv2,lee2023AlignT2I,li2023AGIQA-3K}.
Some pioneering works target solving this problem by annotating a dataset with human preferences, \textit{e.g.}, Pick-a-pic~\cite{kirstain2023Pick-a-pic}, ImageReward~\cite{xu2023ImageReward} and HPD~\cite{wu2023HPSv1,wu2023HPSv2}.
Language-image models, \textit{e.g.} CLIP~\cite{radford2021CLIP} and BLIP~\cite{li2022blip}, are then fine-tuned on the resulting annotated data.
As such, the fine-tuned models represent a data-driven approach to modelling human preferences.
However, the annotation process for capturing human preferences is highly labor- and cost-intensive.
Thus, in this paper, we adopt off-the-shelf image preference models to improve the quality of generated videos.

\noindent\textbf{Learning from human feedback.}
Learning from human feedback was first studied in the context of reinforcement learning and agent alignment~\cite{christiano2017deep_RLHF,leike2018scalable_agent_align} and later in large language models~\cite{ouyang2022InstructGPT,stiennon2020summerize_HF}, enabling them to generate helpful, honest and harmless textual outputs. 
This goal of learning from human feedback is also desirable in visual content generation.
In image generation and editing, a key objective is to align generated images with the prompt~\cite{black2023DDPO,fan2023DPOK,prabhudesai2023AlignProp,clark2023DRaFT,von2023FABRIC,zhang2023HIVE_feedback_edit}, preventing surprising and toxic results~\cite{xu2023ImageReward,wu2023HPSv1,lee2023AlignT2I}.
\citeauthor{lee2023AlignT2I} utilizes reward-weighted regression on a manually collected dataset, aiming to mitigate misalignment with respect to factors such as count, color and background. 
DDPO~\cite{black2023DDPO} and DPOK~\cite{fan2023DPOK} propose to use policy gradients on a multi-step diffusion model~\cite{fan2023DDPM_shortcut_finetuning}, demonstrating improved rewards of aesthetic quality, image-text alignment, compressibility, \textit{etc.}.
DRaFT~\cite{clark2023DRaFT} and AlignProp~\cite{prabhudesai2023AlignProp} achieve feedback optimization by back-propagating the gradients of a differentiable reward function through the sampling procedure via gradient checkpointing~\cite{gruslys2016gradient_checkpointing}.
However, learning from human feedback for video diffusion models remains under-explored owing to its prohibitive cost.
\method aims to fill this gap, providing a solution for more efficient reward fine-tuning.

%% file: sec/03-method.tex
\section{Methodology} \label{sec:method}
In this section, we commence with preliminaries.
Next, we delve into the details of \method, which encompasses:
\textbf{1)} A reformulation of reward fine-tuning as editing that ensures computational efficiency and efficacy.
\textbf{2)} Segmental Video Reward (SegVR) and Temporally Attenuated Reward (TAR) that enable efficient reward fine-tuning with image reward models.

\subsection{Preliminaries} ~\label{sec:preliminary} 
\noindent\textbf{Text-to-video diffusion models.} 
Text-to-video diffusion models aim to map textual input into a distribution representing video data via a reverse diffusion process~\cite{sohl2015Diffusion_model,ho2020denoising_DDPM}. 
These models typically operate in a latent space to handle complex video data~\cite{rombach2022stable_diffusion,wang2023modelscopeT2V,2023videocomposer}.
During pre-training, a sampled video $\bm{x}$ is processed by a fixed encoder~\cite{PatrickEsser2021TamingTF} to derive its latent representation $\bm{z} \in \mathbb{R}^{F \times h \times w \times 3}$, where the video's spatial dimensions are compressed by a factor of $8$.
Next, random noise is injected into the sampled video by the forward diffusion process according to a predetermined schedule ${\{\beta_t\}}_{t=1}^{T}$.
This process can be described as $\bm{z}_t = \sqrt{\bar{\alpha}_t} \bm{z} + \sqrt{1 - \bar{\alpha}_t} \epsilon$, where $\epsilon  \in \mathcal{N}(0,1)$ is random noise with identical dimensions to $\bm{z}$, $\bar{\alpha}_t = \prod_{s=1}^t \alpha_s$ and $\alpha_t = 1 - \beta_t$.
A UNet~\cite{cciccek20163dUNet,ronneberger2015UNet} $\epsilon_{\theta}$ is adopted to perform denoising, enabling the generation of videos through a reverse diffusion process, conditioned on the video caption $\bm{c}$.
Optimisation employs the following reweighted variational bound~\cite{ho2020denoising_DDPM}:
\begin{equation}
    \mathcal{L}(\theta) = \mathbb{E}_{\bm{z}, \epsilon, \bm{c}, t} \left[\| \epsilon - \epsilon_{\theta}(\sqrt{\bar{\alpha}_t} \bm{z} + \sqrt{1 - \bar{\alpha}_t} \epsilon, \bm{c}, t) \|_{2}^{2}\right]
\end{equation}
During inference, we adopt the DDIM sampling~\cite{song2020DDIM} method for realistic video generation.

\noindent\textbf{Reward fine-tuning.}
Reward fine-tuning aims to optimize a pre-trained model to enhance the expected rewards of a reward function $r(\cdot,\cdot)$.
In our case, we target optimizing the parameters $\theta$ of a text-to-video diffusion model to enhance the expected reward of the generated videos given the text distribution:
\begin{equation} ~\label{eqn:reward_obj}
    \mathcal{L}_r (\theta) = \mathbb{E}_{\mathbb{P}(\bm{c})} \mathbb{E}_{\mathbb{P}_{\theta}(x_0|\bm{c})} [- r(\bm{x}_0, \bm{c})]
\end{equation}
where $\bm{x}_0$ is the video generated from the sampled text $\bm{c}$ via the diffusion model through the DDIM sampling chain.
The reward function $r(\cdot,\cdot)$ is typically a pre-trained model to assess the quality of the model output.




\subsection{\textbf{\method}}
In~\cref{fig:pipeline}, we illustrate \method's fine-tuning pipeline and elaborate on the technical contributions below.

\vspace{-.3cm}
\subsubsection{Reward Fine-tuning as Editing}
Reward fine-tuning with diffusion models is costly due to the iterative refinement process during generation using DDIM~\cite{song2020DDIM}.
During generation, initial steps are crucial for shaping coarse, structural aspects of videos, with subsequent steps refining the coarse videos.
Understanding that the essence of reward fine-tuning is not to drastically alter the model's output but to subtly adjust it in line with human preferences, we propose to reinterpret reward fine-tuning as a form of editing~\cite{wu2023tune-a-video,meng2021SDedit,couairon2022Diffedit}. 
This perspective shift allows us to perform partial inference of the DDIM sampling chain, reducing computational demands and easing optimization.

To implement this idea, we first curate a small amount of fine-tuning data from pre-training data for reward fine-tuning.
For each video-text pair $(\bm{x}, \bm{c})$, we acquire the video's latent embedding $\bm{z}$ as stated in~\cref{sec:preliminary}.
We aim to smooth out the video to eliminate undesirable artifacts and distortions~\cite{meng2021SDedit}.
To achieve this, we leverage the diffusion process rather than DDIM inversion~\cite{couairon2022Diffedit,meng2021SDedit} to enable efficient editing.
If we denote the number of DDIM steps as $D$ and the number of pre-training DDPM steps as $T$, we define a mapping $d: \{1, \ldots, D\} \rightarrow \{1, \ldots, T\}$ that maps DDIM step index to the DDPM step index\footnote{For example, if $D=20$ and $T=1000$, then the DDIM step sub-sequence is $\{ 1, 51, \ldots, 901, 951\}$, \textit{i.e.}, $d(2) = 51$.}, formulated as:
\begin{equation}
    d(i) = \smash{\frac{T}{D}} \cdot (i-1) + 1
\end{equation}
Given the noise level $\tau$, the targeted diffusion step $t_{\rm noi}$ for injecting noise is formulated as:
\begin{equation}
    t_{\rm noi} = d(\tau \cdot D)
\end{equation}
This allows us to obtain the starting point for reward fine-tuning via the diffusion process:
\begin{equation}
    \bm{z}_{t_{\rm noi}} = \sqrt{\bar{\alpha}_{t_{\rm noi}}} \bm{z} + \sqrt{1 - \bar{\alpha}_{t_{\rm noi}}} \epsilon
\end{equation}
Based on $\bm{z}_{t_{\rm noi}}$, we can perform $\tau \times D$ steps of DDIM sampling~\cite{song2020DDIM} along the DDIM sub-sequence to obtain the edited result $\bm{z}_0$, which consumes $\tau$ of the computation of the full sampling chain.
Utilizing the decoder~\cite{PatrickEsser2021TamingTF}, we decode $\bm{z}_0$ in the latent space to $\bm{x}_0$ in the RGB space.

\vspace{-.3cm}
\subsubsection{Reward Fine-tuning with Image Reward Models}
Since curating large datasets to capture human preferences for training video reward models is prohibitively expensive, we resort to off-the-shelf image reward models $r(\cdot,\cdot)$, \textit{e.g.}, HPSv2~\cite{wu2023HPSv2}.
HPSv2 is trained on 430$\rm k$ pairs of images, which are annotated by humans for text-image alignment and image quality.
Given that videos are natural extensions of images, we posit that these human preferences are also applicable to videos.
However, initial experiments with applying dense reward fine-tuning produced degraded motion continuity.
Taking inspiration from temporal segment networks~\cite{wang2016TSN}, given a video $\bm{x}_0 \in \mathbb{R}^{F \times H \times W \times 3}$ generated from its caption $\bm{c}$, we evenly divide it into $S$ segments.
Within each segment, we perform random frame sampling, obtaining a sparse set capturing the essence of the video $\bm{x}_0^{g} = \{ \bm{x}_0^{g(1)},\ldots,\bm{x}_0^{g(S)} \}$.
Here, $g(i) = {\rm Uniform}\left( (i-1) \cdot \frac{F}{S}, i \cdot \frac{F}{S} - 1 \right)$ denotes a uniform sampling of index within $i$th segment.
Utilizing $r(\cdot,\cdot)$, we compute the reward score $R$ with respect to $\bm{x}_0$ as follows:
\begin{equation}
    R = {\rm Agg}_i [r(\bm{x}_0^{g(i)}, \bm{c})],\quad i = 1,\ldots,S
\end{equation}
where ${\rm Agg}_i$ denotes the aggregation function along index $i$.
To consider the impact of all frames in $\bm{x}_0^{g}$, an intuitive implementation of ${\rm Agg}_i$ is the $\rm mean$ function.

However, the simple aggregation function leads to noticeable visual artifacts in the generated videos, such as structural twitching and color jittering.
This issue arises because the $\rm mean$ function places equal weight on all frames, disregarding the inherent dynamic nature of videos where the reward scores of frames can vary throughout the sequence.
To address this, we introduce TAR that strategically emphasizes central frames, with the emphasis tapering off towards the peripheral frames, thereby avoiding uniformly optimizing all frames' reward scores to be equally high.
We define the temporally attenuated coefficient as:
\begin{equation}
    f_i = {\rm e}^{-\lambda_{\rm tar} |g(i) - \frac{F}{2}|}
\end{equation}
where $\lambda_{\rm tar}$ controls the degree of the attenuating rate.
We set $\lambda_{\rm tar} = 1$ by default.
Incorporating this coefficient, we rewrite the reward score $R$:
\begin{equation}
    R = \frac{1}{S} \sum\nolimits_{i = 1}^{S}{f_i \cdot r(\bm{x}_0^{g(i)}, \bm{c})}
\end{equation}
The optimization objective in~\cref{eqn:reward_obj} can be rewritten as:
\begin{equation}
    \mathcal{L}_r (\theta) = \mathbb{E}_{\mathbb{P}(\bm{c})} \mathbb{E}_{\mathbb{P}_{\theta}(x_0|\bm{c})} [- R]
\end{equation}

\begin{figure*}[t]
\centering
    \includegraphics[width=1.\textwidth]{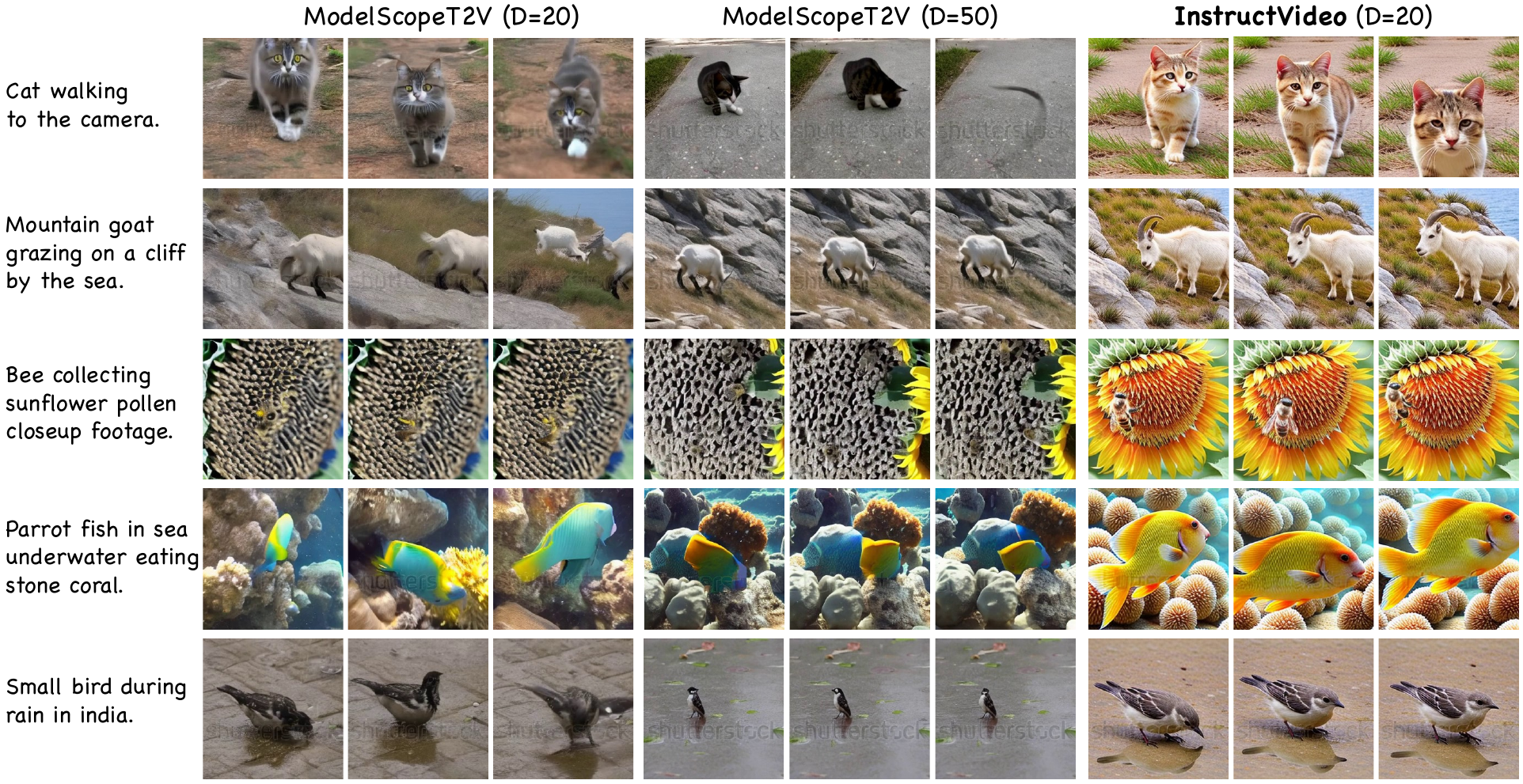}
    \vspace{-0.6cm}
    \caption{\small 
        \textbf{Comparion of \method with the base model ModelScopeT2V}.
        ModelScopeT2V utilizes 20 and 50 DDIM steps.
    }
\vspace{-0.3cm}
\label{fig:comparison_with_ModelScopeT2V}
\end{figure*}

\subsection{Reward Fine-tuning and Inference}
\noindent\textbf{Data preparation and evaluation metric.}
We follow DDPO~\cite{black2023DDPO} to experiment on prompts describing 45 animal species.
In contrast to DDPO, since \method relies on video-text data, we select video-text pairs as the \textit{fine-tuning data} from the base model's pre-training dataset, \textit{i.e.}, WebVid10M, ensuring that no extra data is introduced.
It is worth noting that we \textbf{do not apply} any quality filtering method to ensure that the selected videos are of high quality.
Specifically, we select about 20 video-text pairs for each animal species.
To evaluate the model's ability to optimize the reward scores, we also collect \textit{evaluation data} comprising about 6 prompts for each animal.
We use HPSv2 score to measure the optimization performance of reward fine-tuning on the first frames of all segments. 

\noindent\textbf{Reward fine-tuning.} 
We adopt the publicly available text-to-video diffusion model ModelScopeT2V~\cite{wang2023modelscopeT2V} as our base model.
ModelScopeT2V is trained on WebVid10M~\cite{Bain21Webvid} with $T=1000$ and is able to generate videos of $16 \times 256 \times 256$ resolution, which we divide into $S=4$ segments. 
By default, we adopt the differentiable HPSv2~\cite{wu2023HPSv2} as the reward model and perform 20-step DDIM inference, \textit{i.e.}, $D = 20$.
Classifier-free guidance~\cite{ho2022classifier} is adopted by default.
Directly back-propagating the reward loss to the diffusion models can be computationally intensive and risks catastrophic forgetting~\cite{kirkpatrick2017overcomingforgetting,Feng2022PLwF,feng2022overcoming}. 
To circumvent these issues, we incorporate LoRA~\cite{hu2021lora} by default.
To further accelerate fine-tuning, we truncate the gradient to only back-propagate the last DDIM sampling step following~\cite{clark2023DRaFT}.
Experiments are conducted on 4 NVIDIA A100s, with the batch size set to $8$ and the learning rate set to $1 \times 10^{-5}$.
To strike a cost-performance balance, we fine-tune \method with default parameters for 20$\rm k$ steps if not otherwise stated.

\noindent\textbf{Inference.}
After reward fine-tuning, we merge the LoRA weights into the ModelScopeT2V parameters to ensure that \method's inference cost is identical to ModelScopeT2V~\cite{hu2021lora}.
For text-to-video generation, \method uses 20-step DDIM inference.

%% file: sec/04-exp.tex
\section{Experiments}\label{sec:exp}



\subsection{Effectiveness of \textbf{\method}}

\begin{figure*}[t]
\centering
    \includegraphics[width=1.\textwidth]{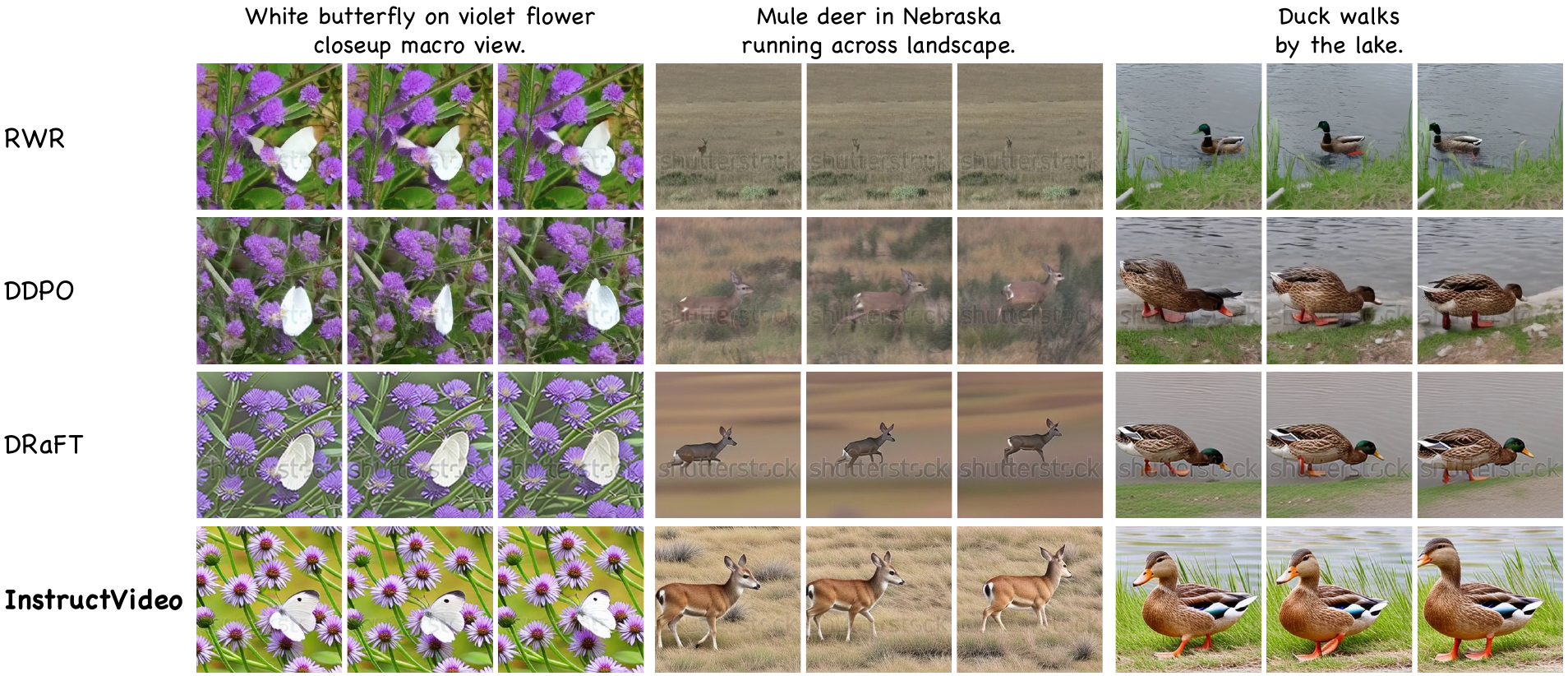}
    \vspace{-0.6cm}
    \caption{\small 
        \textbf{Comparison of \method with other reward fine-tuning methods}.
        We set $D=20$ for all methods.
    }
\vspace{-0.3cm}
\label{fig:comparing-with-other-reward-finetuning-examples}
\end{figure*}

\noindent\textbf{Comparison with the base model ModelScopeT2V.}
To verify the efficacy of \method, we compare it with ModelScopeT2V~\cite{wang2023modelscopeT2V} utilizing $20$ and even $50$ DDIM steps in~\cref{fig:comparison_with_ModelScopeT2V}.
Examining the examples, we observe that the quality of videos generated by \method consistently outperforms the base model by a margin.
Specifically, notable enhancements include 
\textbf{1)} clearer and more coherent structures and scenes even if the animal is moving, exemplified by the walking cat and the swimming fish;
\textbf{2)} more appealing coloration, exemplified by the sunflower, the bee and the mountain goat;
\textbf{3)} an enhanced delineation of scene details, exemplified by the rock and grass on the cliff, and the texture of all the animals;
and \textbf{4)} improved video-text alignment, exemplified by the distinct portrayal of sunflowers and the bird's reflections on the water.
Remarkably, these advancements are achieved without compromising motion fluidity and the resultant videos can often surpass the video quality of the WebVid10M dataset.
Notably, \method even attenuates watermarks present in WebVid10M.
These qualitative leaps, consistently favored by human annotators, are attributed to the reward fine-tuning process, which effectively refines the video diffusion model.



\begin{figure*}[t]
\centering
    \includegraphics[width=1.\textwidth]{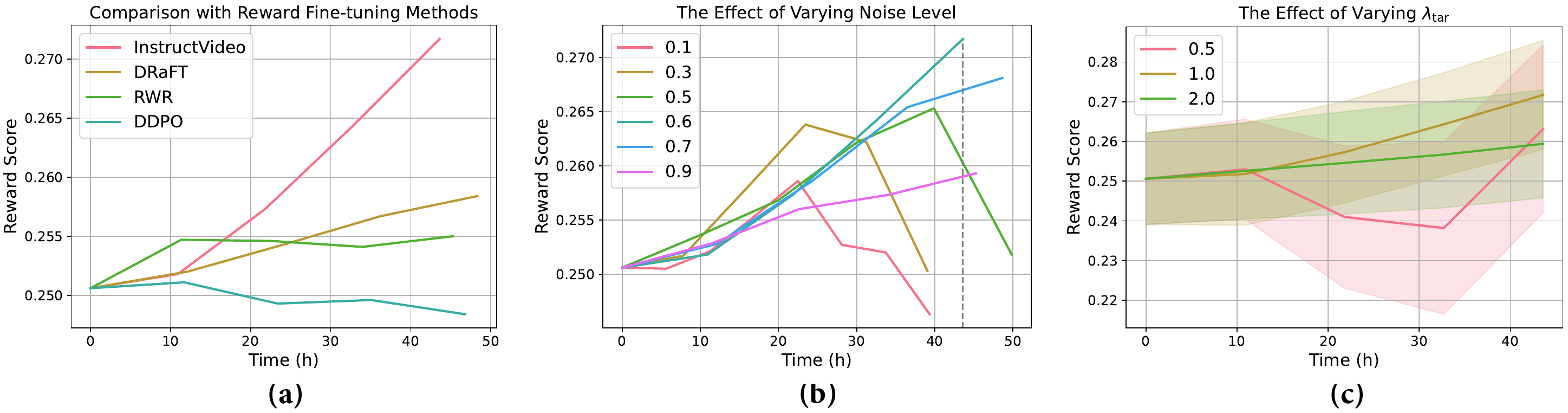}
    \vspace{-0.6cm}
    \caption{\small 
        \textbf{(a) Comparison with other reward fine-tuning methods}.
        \textbf{(b) The effect of varying noise level $\tau$.}
        The vertical dashed line indicates 20$\rm k$ steps of optimization for $\tau = 0.6$.
        \textbf{(c) The effect of varying $\lambda_{\rm tar}$.}
    }
\vspace{-0.3cm}
\label{fig:combined_curves}
\end{figure*}

\noindent\textbf{Comparison with other reward fine-tuning methods.}
This aims to validate the efficacy of reward fine-tuning conceptualized as an editing process.
We compare with other representative reward fine-tuning methods, including policy gradient algorithm, DDPO~\cite{black2023DDPO}, reward-weighted regression, RWR~\cite{lee2023AlignT2I} and direct reward back-propagation method, DRaFT~\cite{clark2023DRaFT}.
For DRaFT, we adopt DRaFT-1 for efficient fine-tuning.
SegVR and TAR are employed for all compared methods to standardize reward signals.
The comparative analysis on the evaluation set, presented in~\cref{fig:combined_curves}(a), leads to two findings:
\textbf{1)} Both RWR and DDPO exhibit a performance plateau after about 11 hours of fine-tuning, with further optimization failing to enhance or even deteriorating performance.
\textbf{2)} Direct reward back-propagation methods, including \method and DRaFT, initially lag during the first 11 hours but subsequently demonstrate fine-tuning efficiency, especially \method.
To further validate the efficacy of our method, we provide visual comparisons in~\cref{fig:comparing-with-other-reward-finetuning-examples}, where we adopt the optimal fine-tuned checkpoint for each method in~\cref{fig:combined_curves}(a).
The examples reflect \method’s superiority, evident in:
\textbf{1)} the clarity and coherence of structural and scenic elements,
\textbf{2)} the vibrancy of colors,
\textbf{3)} the precision in depicting intricacies,
and \textbf{4)} enhanced video-text alignment.

\begin{figure*}[t]
\centering
    \includegraphics[width=1.\textwidth]{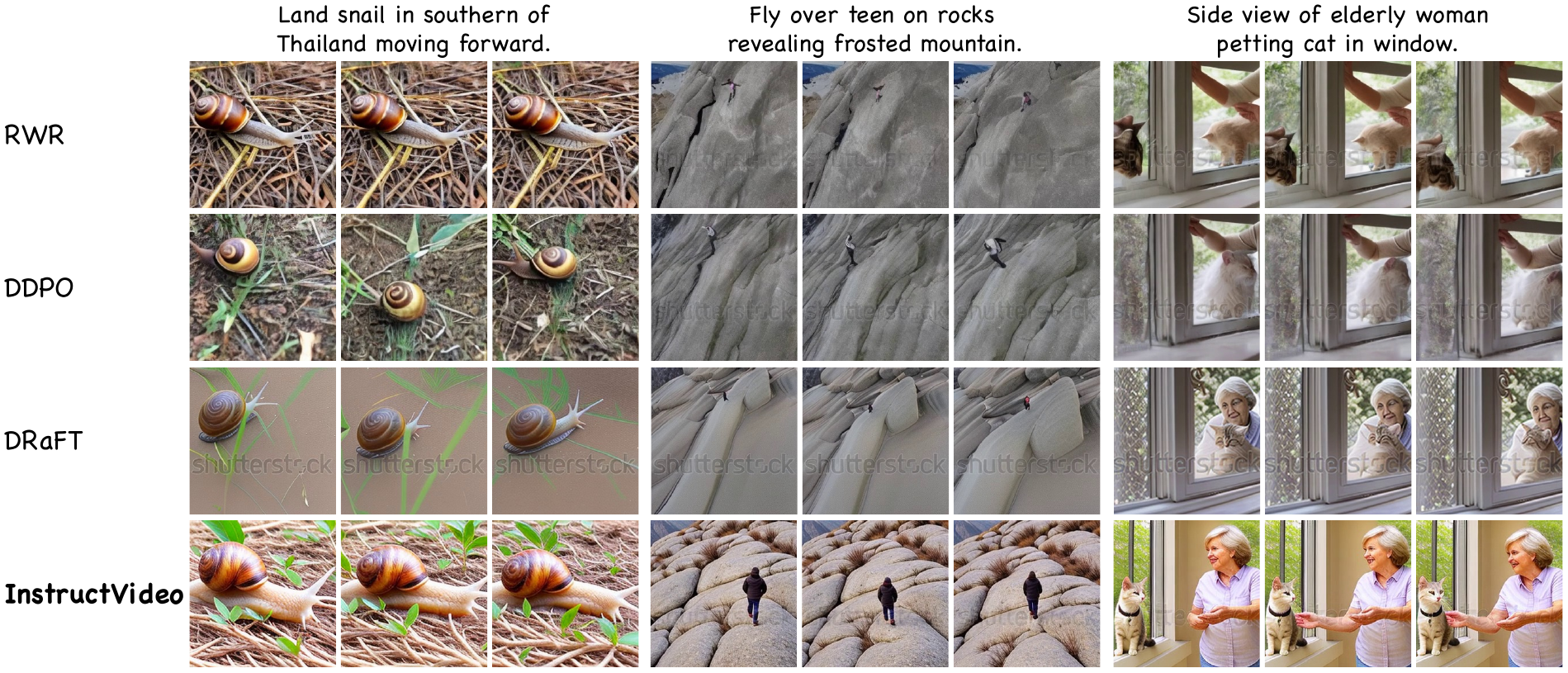}
    \vspace{-0.6cm}
    \caption{\small 
        \textbf{Comparison of \method's generalization capabilities with other methods}.
        We set $D=20$ for all methods.
    }
\vspace{-0.4cm}
\label{fig:generalization-to-new-prompts-examples}
\end{figure*}


\noindent\textbf{Generalization to unseen text prompts.}
We assessed the model's generalization capabilities using two distinct sets of prompts: 
\textbf{1)} those describing new animals and 
\textbf{2)} those related to non-animals, none of which are present in the fine-tuning data.
For this purpose, we curate about 4 prompts each for 6 new animal species following~\cite{prabhudesai2023AlignProp} and 46 prompts for non-animals.
Our comparative analysis of \method, the base model ModelScopeT2V and other reward fine-tuning methods is presented in~\cref{tab:generalization_to_new_prompts}.
It is worth noting that during fine-tuning, SegVR and TAR are adopted by default to provide reward signals.
Our observations are threefold:
\textbf{1)} Increasing the number of DDIM steps enhances the video quality.
\textbf{2)} Compared to the base model ModelScopeT2V in the second row, all methods improve reward scores for these unseen prompts.
\textbf{3)} Among all methods, \method outperforms other alternatives, affirming its superior generalization capabilities.
To further demonstrate this, we provide visual comparisons in~\cref{fig:generalization-to-new-prompts-examples}, featuring a set of unseen animal species, various sceneries, and human figures.
The presented examples display an enhanced quality and exhibit an improvement in video-text alignment.


\begin{table}[t]
  \small
  \setlength{\tabcolsep}{3pt}
  \centering
    \begin{tabular}{c|ccc}
    \textbf{Method} & \textbf{In-domain} & \textbf{New Animals} & \textbf{Non-animals} \\
    \shline
    ModelScopeT2V\textsuperscript{\dag} & 0.2542 \tiny{± 0.0122} & 0.2541 \tiny{± 0.0109} & 0.2610 \tiny{± 0.0158} \\
    ModelScopeT2V & 0.2506 \tiny{± 0.0155} & 0.2502 \tiny{± 0.0138} & 0.2557 \tiny{± 0.0177} \\
    DDPO~\cite{black2023DDPO} & 0.2511 \tiny{± 0.0114} & 0.2524 \tiny{± 0.0112} & 0.2564 \tiny{± 0.0171} \\
    RWR~\cite{lee2023AlignT2I} & 0.2550 \tiny{± 0.0166} & 0.2517 \tiny{± 0.0101} & 0.2625 \tiny{± 0.0146} \\
    DRaFT~\cite{clark2023DRaFT} & 0.2584 \tiny{± 0.0123} & 0.2561 \tiny{± 0.0098} & 0.2644 \tiny{± 0.0174} \\
    \hline
    \textbf{\method} & \textbf{0.2717 \tiny{± 0.0137}} & \textbf{0.2645 \tiny{± 0.0125}} & \textbf{0.2682 \tiny{± 0.0202}} \\
    \end{tabular}%
    \vspace{-0.2cm}
    \caption{
        \textbf{Generalization to unseen text prompts}.
        \dag \ denotes the model utilizes $D=50$ while others adopt $D=20$.
        `In-domain' denotes in-domain animal prompts from the evaluation data.
    }
    \vspace{-0.4cm}
  \label{tab:generalization_to_new_prompts}%
\end{table}%




\noindent\textbf{User study.}
To further qualitatively compare the videos generated by \method and other methods, we conduct a user study comparing our methods with the base model ModelScopeT2V and the reward fine-tuned model DRaFT in~\cref{tab:user_study}.
We recruited two participants who have related research experience in generative models to assess the quality of videos in terms of video quality and video-text alignment.
To simplify the annotation process, participants were presented with pairs of videos and asked to identify which video was superior or if both were of equal quality.
To ensure a comprehensive comparison, we chose 60 prompts from the 45 fine-tuning animal species, 20 prompts from the 6 new animal species and 20 prompts describing non-animals.
More details are presented in the Appendix.
We observe that our method consistently outperforms other methods.
Specifically, improvements in video quality, a noted shortcoming of the base model, are more pronounced than improvements in video-text alignment.

\begin{table}[t]
  \small
  \setlength{\tabcolsep}{2pt}
  \centering
    \begin{tabular}{c|ccc|ccc}
    \textbf{\method} & \multicolumn{3}{c|}{\textbf{Quality}} & \multicolumn{3}{c}{\textbf{Alignment}} \\
    \textbf{vs} & Ours  & Tie   & Other & Ours  & Tie  & Other \\
    \shline
    ModelScopeT2V & 75.5\%  & 15.5\%   & 9.0\%  & 28.5\%  & 57.0\%  & 14.5\% \\
    DRaFT & 76.0\%  & 10.5\%  & 13.5\%  & 30.0\%  & 51.5\% & 18.5\% \\
    \end{tabular}
    \vspace{-0.2cm}
    \caption{
    \small
    \textbf{User study.}
    `Tie' indicates instances where annotators think two videos are of comparable quality.
    `Quality' and `Alignment' represent video quality and video-text alignment.
    }
    \vspace{-0.3cm}
  \label{tab:user_study}
\end{table}

\subsection{Ablation Study}
\noindent\textbf{The effect of varying noise level $\tau$.}
To determine the optimal choice for noise level $\tau$, we vary its value and evaluate its impact on reward scores using the evaluation data.
We illustrate the results in~\cref{fig:combined_curves}(b).
An increase in $\tau$ from $0.1$ to $0.5$ correlates with a progressive enhancement in the highest reward scores achieved by \method. 
However, excessively prolonged fine-tuning precipitates a sharp decline in generative performance.
This phenomenon can be attributed to the limited edited space available to the model at lower noise levels, which constricts its ability to find the optimal output space as directed by the reward scores.
When we further increase $\tau$ from $0.6$ to $0.9$, we observe that the reward score enhancement per hour becomes minor, suggesting challenges associated with generating from an extended sampling chain.
Optimally, a noise level of $\tau = 0.6$ strikes a balance, providing a feasible starting point for editing that still allows for a substantial exploration of the edited space.
After 20$\rm k$ steps, more optimization leads to over-optimization~\cite{prabhudesai2023AlignProp}, meaning that further steps can degrade the visual quality of the output despite potential increases in the reward score. 
Thus, we finalize on $\tau = 0.6$ with 20$\rm k$ steps of fine-tuning.



\noindent\textbf{The effect of varying $\lambda_{\rm tar}$.}
To determine the optimal choice for $\lambda_{\rm tar}$, we vary its value and evaluate its impact.
We illustrate the results in~\cref{fig:combined_curves}(c).
A relatively high value $\lambda_{\rm tar}$, such as $2.0$, results in $f_i$ decaying exponentially faster towards those border frames, thus providing diminished reward signals.
A relatively low value $\lambda_{\rm tar}$, such as $0.5$, leads to $f_i$ decaying more gently towards those border frames, thus strengthening the reward signals.
This equalized weighting across frames can destabilize fine-tuning, leading to a precipitous decline in reward scores.
Subsequent increases in scores do not necessarily indicate improved video quality but indicate quality degradation, as revealed by the rising variance.
Thus, an appropriate coefficient to ensure stable fine-tuning is imperative and we finalize on $\lambda_{\rm tar} = 1.0$.

\begin{figure}[t]
\centering
    \includegraphics[width=.48\textwidth]{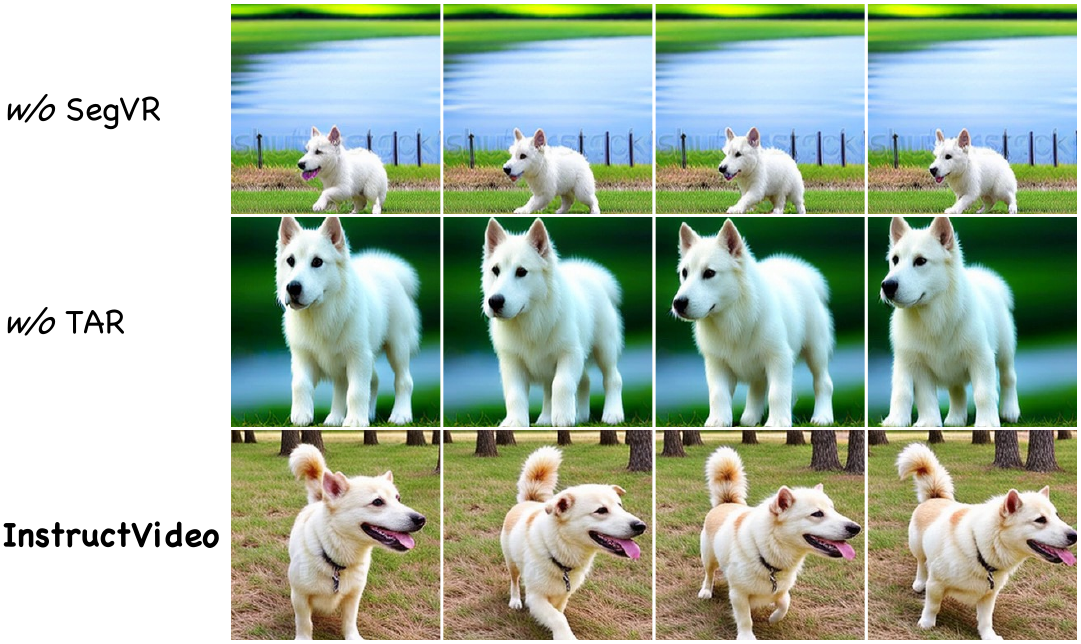}
    \vspace{-0.6cm}
    \caption{\small 
        \textbf{Ablation study on SegVR and TAR}.
        This video shows a white dog walking in the park in slow motion.
    }
\vspace{-0.3cm}
\label{fig:ablation-on-SegVR-TAR}
\end{figure}

\noindent\textbf{Ablation on SegVR and TAR.}
To qualitatively verify the efficacy of SegVR and TAR, we present illustrative results in~\cref{fig:ablation-on-SegVR-TAR}.
Removing either SegVR or TAR results in a noticeable reduction in temporal modeling capabilities. 
This suggests that overly dense or excessively strong reward signals can lead to generation collapse.
The degradation of modeling temporal dynamics often leads to the degraded quality of the individual frames due to the intertwined nature of spatial and temporal parameters.
These observations underscore the critical roles of SegVR and TAR in maintaining fine-tuning stability.

\subsection{Further Analysis}

\begin{figure}[t]
\centering
    \includegraphics[width=.48\textwidth]{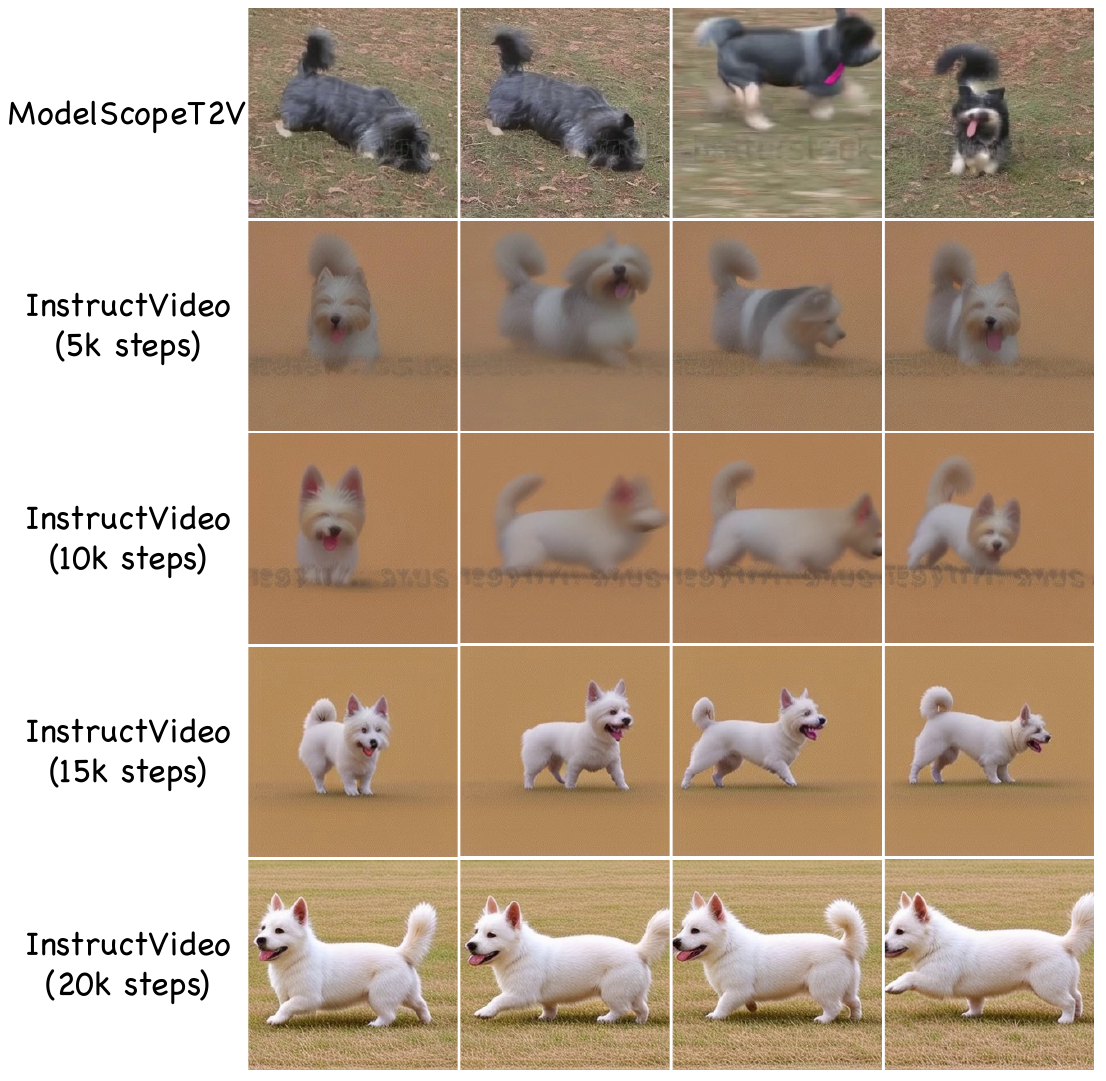}
    \vspace{-0.6cm}
    \caption{\small 
        \textbf{The evolution of generated videos} during fine-tuning.
        The video shows a bobtail dog walking.
    }
\vspace{-0.3cm}
\label{fig:videos_evolve}
\end{figure}

\noindent\textbf{The evolution of the generated videos during reward fine-tuning.}
To elucidate how the reward fine-tuning works, we present a visual progression in~\cref{fig:videos_evolve}.
The top row depicts a video generated without fine-tuning.
All the frames in this video exhibit a lack of the dog’s fur texture.
Moreover, a notable blurriness characterizes the third frame due to sudden and unanticipated motion, while the fourth frame suffers from a loss of facial clarity.
As reward fine-tuning proceeds, we observe a noticeable enhancement in terms of all aspects mentioned above.  
Surprisingly, watermarks, which are consistently present across the dataset, also gradually fade. 
The resultant video is full of clear details and aesthetically pleasing coloration.

\begin{figure}[t]
\centering
    \includegraphics[width=.4\textwidth]{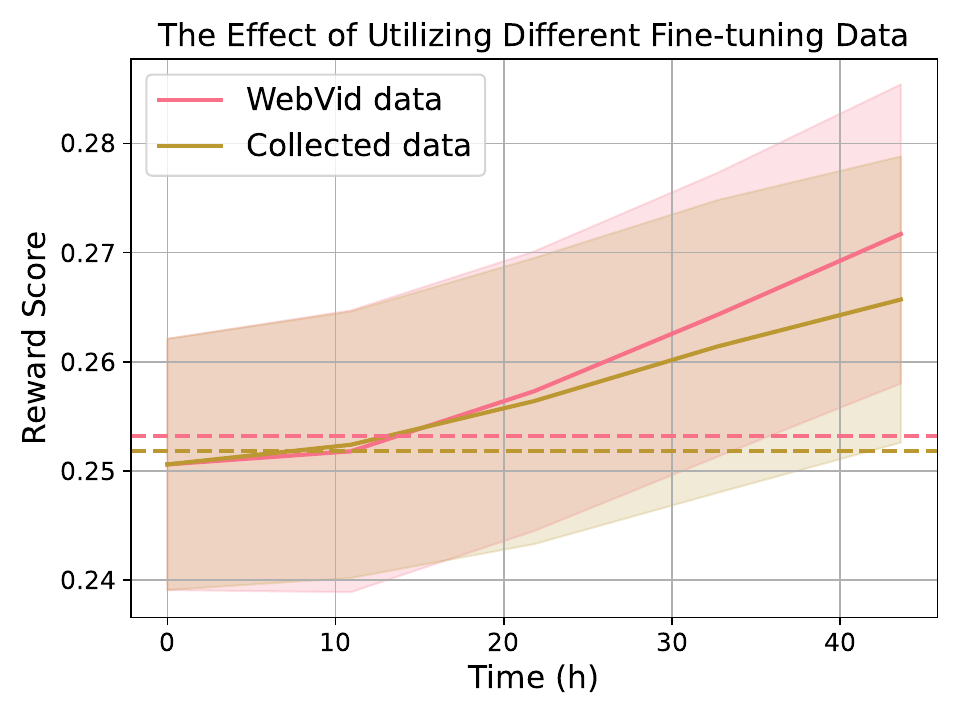}
    \vspace{-0.4cm}
    \caption{\small 
        \textbf{The effect of utilizing different fine-tuning data.}
        The colored horizontal dashed lines denote the reward scores of different fine-tuning data, matched by the color of the respective curve.
    }
\vspace{-0.3cm}
\label{fig:diff_finetuning_data}
\end{figure}

\noindent\textbf{Impact of fine-tuning data quality on the fine-tuning results.}
To investigate this, we self-collect a dataset comprising an equivalent number of video-caption pairs for 45 animal species, which are employed for fine-tuning.
The results are illustrated in~\cref{fig:diff_finetuning_data}.
We employ horizontal dashed lines to indicate the quality of different data, inferred from reward scores.
While the variance of the generated videos are comparable for two kinds of fine-tuning data, the employment of higher-quality data, \textit{i.e.}, WebVid10M, yields superior average reward scores compared to that obtained using the lower-quality counterpart.
This suggests that superior fine-tuning data can facilitate reward fine-tuning.


\noindent\textbf{Constraints of fine-tuning data on resultant video quality.}
\cref{fig:diff_finetuning_data} showcases that \method is capable of generating videos that achieve reward scores significantly exceeding those of the fine-tuning data itself, as denoted by the horizontal dashed lines. 
This observation leads us to conclude that the quality of the fine-tuning data does not impose a ceiling on the potential quality of the fine-tuned results. 
Our fine-tuning pipeline has the propensity to surpass the initial data quality, thus facilitating the generation of videos with substantially enhanced reward scores.


%% file: sec/05-conclusion.tex
\section{Conslusion}
In this paper, we introduce \method, a method that pioneers instructing video diffusion models with human feedback by reward fine-tuning.
We recast reward fine-tuning as an editing process that mitigates computational burden and enhances fine-tuning efficiency.
We resort to image reward models to provide human feedback on generated videos and propose SegVR and TAR to ensure effective fine-tuning.
Extensive experiments validate that \method not only elevates visual quality but also maintains robust generalization capabilities.

%% file: sec/06-supp.tex
\clearpage
\setcounter{page}{1}
\maketitlesupplementary
\appendix
\renewcommand{\thefigure}{A.\arabic{figure}}
\setcounter{figure}{0}
\renewcommand{\thetable}{A.\arabic{table}}
\setcounter{table}{0}

In this Appendix, we first detail on the potential societal impact (\cref{app:societal_impact}), and limitations and potential future work  (\cref{app:limitations}).
Subsequently, we provide additional details about implementing LoRA (\cref{app:more_about_lora}) and user study (\cref{app:more_about_user_study}).
Moreover, we provide more visualization results to demonstrate the efficacy of \method (\cref{more_visualization}).
Finally, we provide experiments to validate the efficacy of generation with 50-step DDIM inference (\cref{app:50-step_generation}), reward fine-tuning with 50-step DDIM inference (\cref{app:50-step_reward_fine-tuning}) and \method's adaptation to other reward functions (\cref{app:adaptation_to_other_reward_functions}).

\section{Potential Societal Impact}\label{app:societal_impact}
\method, as the pioneering effort in instructing video diffusion models with human feedback, prioritizes users' preferences for AI-generated content.
We conducted this research, motivated by the varied quality of generated videos induced by the varied quality of the curated web-scale datasets.
Pre-training models on such unfiltered data can lead to outputs that deviate from human preferences.
In the context of the broader research community, we advocate that video generation systems, akin to other generative models like language models~\cite{openai2023gpt4,ouyang2022InstructGPT}, should prioritize ethical considerations and human values.

Moreover, conventional video generation systems might not always resonate with all users in terms of aesthetic style and often struggle with accurately reflecting textual prompts.
\method steps in as a human-centered technology, efficiently addressing these issues in a data- and computation-efficient way and opening up possibilities for commercial applications, particularly in sectors like education and entertainment.

However, as \method primarily targets research, aiming at investigating the practicality of aligning video diffusion models with human preferences, its deployment to any circumstance beyond research should be approached with thorough oversight and evaluation to ensure responsible and ethical use.

\section{Limitation and Future work}\label{app:limitations}
We recognize that \method, as an initial endeavor in this area, comes with its limitations.
Although we validate the efficacy of image reward models, we anticipate that specialized video reward models capturing human preferences might be even more superior since they evaluate one generated video as a whole.
Additionally, as a common issue mentioned in previous works~\cite{prabhudesai2023AlignProp,dong2023RAFT,black2023DDPO,fan2023DPOK}, reward fine-tuning carries a risk of over-optimization, meaning that excessive optimization steps will result in the degradation of the video quality despite potential increases in the reward score.
Addressing these aspects presents avenues for future research, including the development of a more advanced video reward model and the design of strategic mechanisms to identify and ameliorate over-optimization.




\section{More Details about Implementing LoRA}\label{app:more_about_lora}
To instantiate LoRA~\cite{hu2021lora} for efficient tuning, we adopt the implementation used in Diffusers\footnote{\url{https://github.com/huggingface/diffusers/blob/main/src/diffusers/models/lora.py}}.
Specifically, we configure the intrinsic rank within LoRA to 4 to ensure fast processing.
LoRA modifications are applied to every Transformer~\cite{AttentionAlluNeed} layer within our model, targeting the linear layers responsible for query, key, value, and output projections.
ModelScopeT2V~\cite{wang2023modelscopeT2V} contains 1,347.44M parameters, whereas the additional parameters introduced by adding LoRA amount to only 1.58M – approximately \textbf{0.1\%} of the total ModelScopeT2V parameters.


\section{More Details about User Study}\label{app:more_about_user_study}
In the main paper, we present a user study to demonstrate the effectiveness of \method. This study involves a comparative analysis of videos generated by \method and other methods, focusing on two key aspects: video quality and video-text alignment.
For video quality, we asked annotators to evaluate:
\textbf{1)} The overall visual quality of the videos,
\textbf{2)} Alignment with general human aesthetic preferences, such as pleasing visuals, texture and details, and
\textbf{3)} The smoothness and consistency in terms of structural and color transitions within the video.
Regarding video-text alignment, annotators are tasked with determining the extent to which the generated videos accurately and clearly represent the content of the provided text prompts. 
This assessment included evaluating the depiction of entities, attributes, relationships, and motions as described in the prompts.
To simplify the evaluation process, annotators are asked to perform pairwise comparisons between videos, thereby streamlining their task to direct contrasts rather than isolated assessments.


\section{More Visualization Results}\label{more_visualization}
\begin{figure*}[t]
\centering
    \includegraphics[width=1.\textwidth]{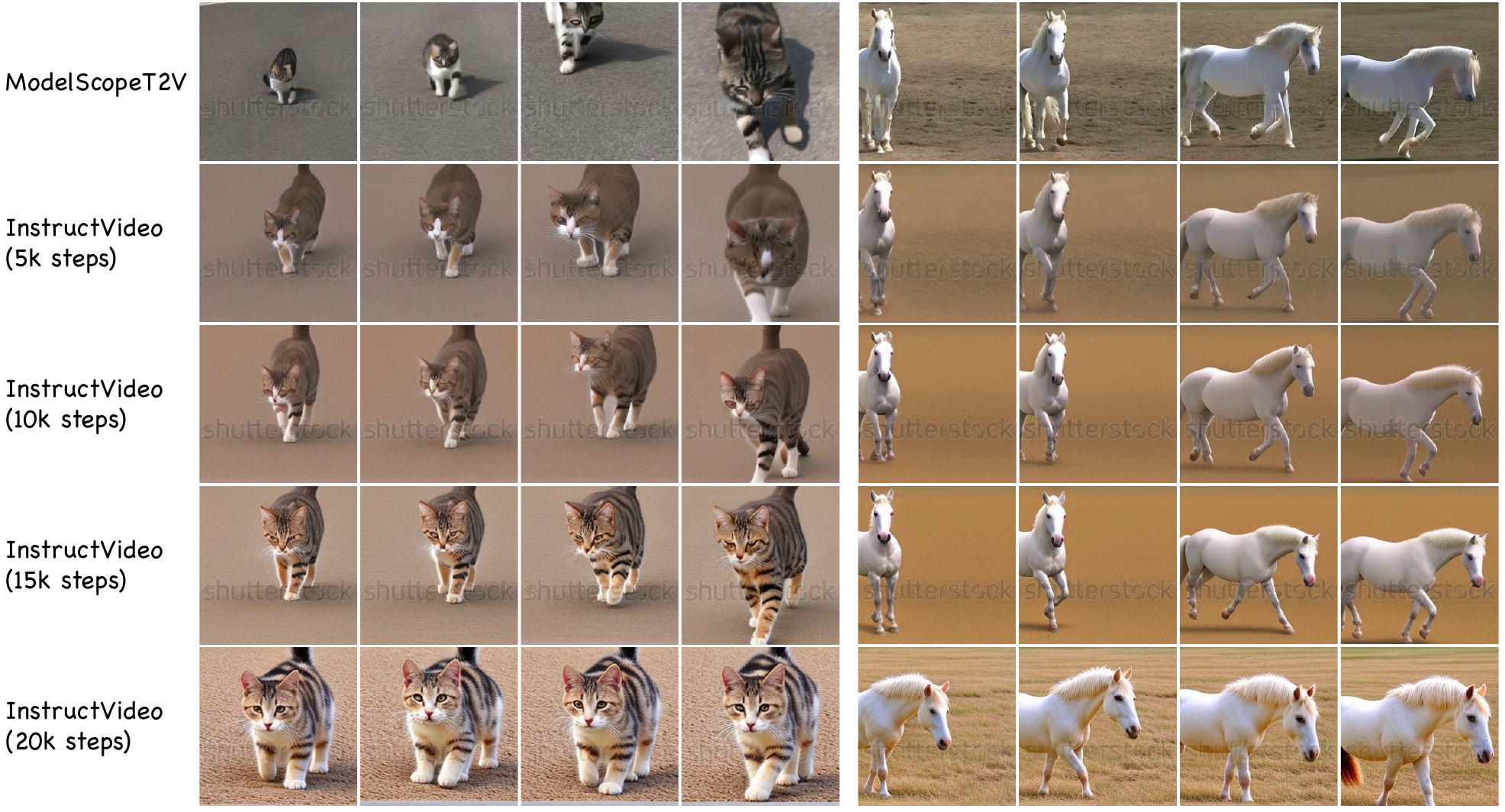}
    \vspace{-0.3cm}
    \caption{\small 
        More examples showing \textbf{the evolution of generated videos} during fine-tuning.
    }
\vspace{-0.3cm}
\label{fig:videos_evolve_supp}
\end{figure*}

\begin{figure*}[t]
\centering
    \includegraphics[width=1.\textwidth]{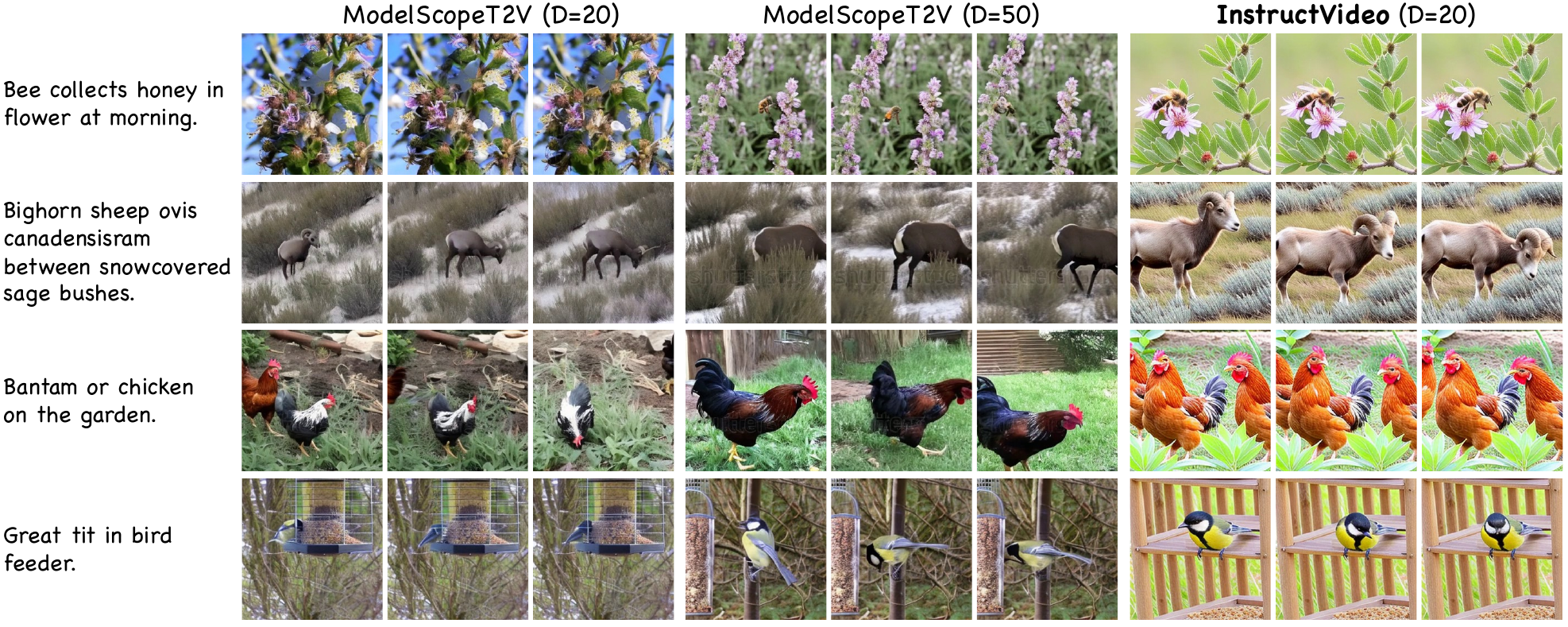}
    \vspace{-0.3cm}
    \caption{\small 
        More examples showing the \textbf{comparion of \method with the base model ModelScopeT2V}.
        ModelScopeT2V utilizes 20 and 50 DDIM steps.
    }
\vspace{-0.3cm}
\label{fig:comparison_with_ModelScopeT2V_supp}
\end{figure*}

\begin{figure*}[t]
\centering
    \includegraphics[width=1.\textwidth]{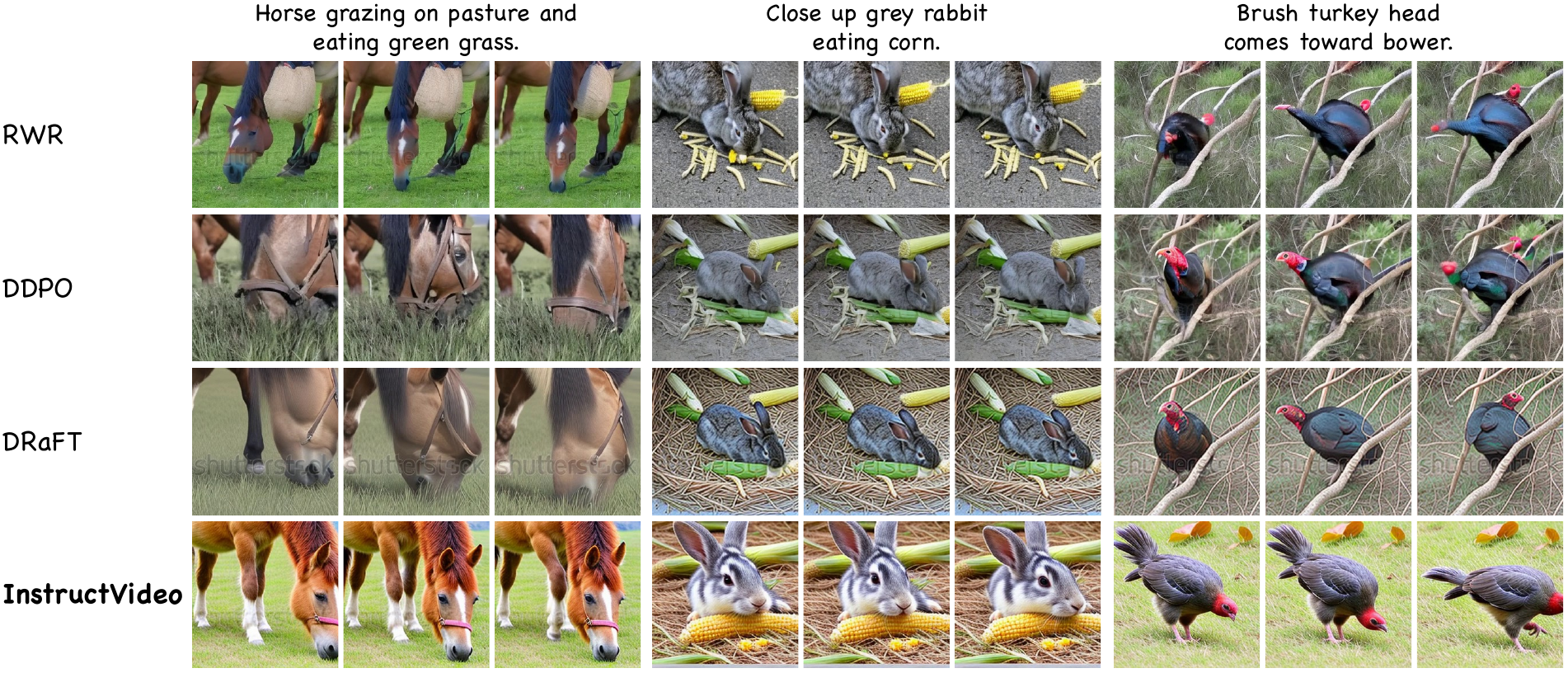}
    \vspace{-0.3cm}
    \caption{\small 
        More examples showing the \textbf{comparison of \method with other reward fine-tuning methods}.
        We set $D=20$ for all methods.
    }
\vspace{-0.3cm}
\label{fig:comparing-with-other-reward-finetuning-examples-supp}
\end{figure*}

\begin{figure*}[t]
\centering
    \includegraphics[width=1.\textwidth]{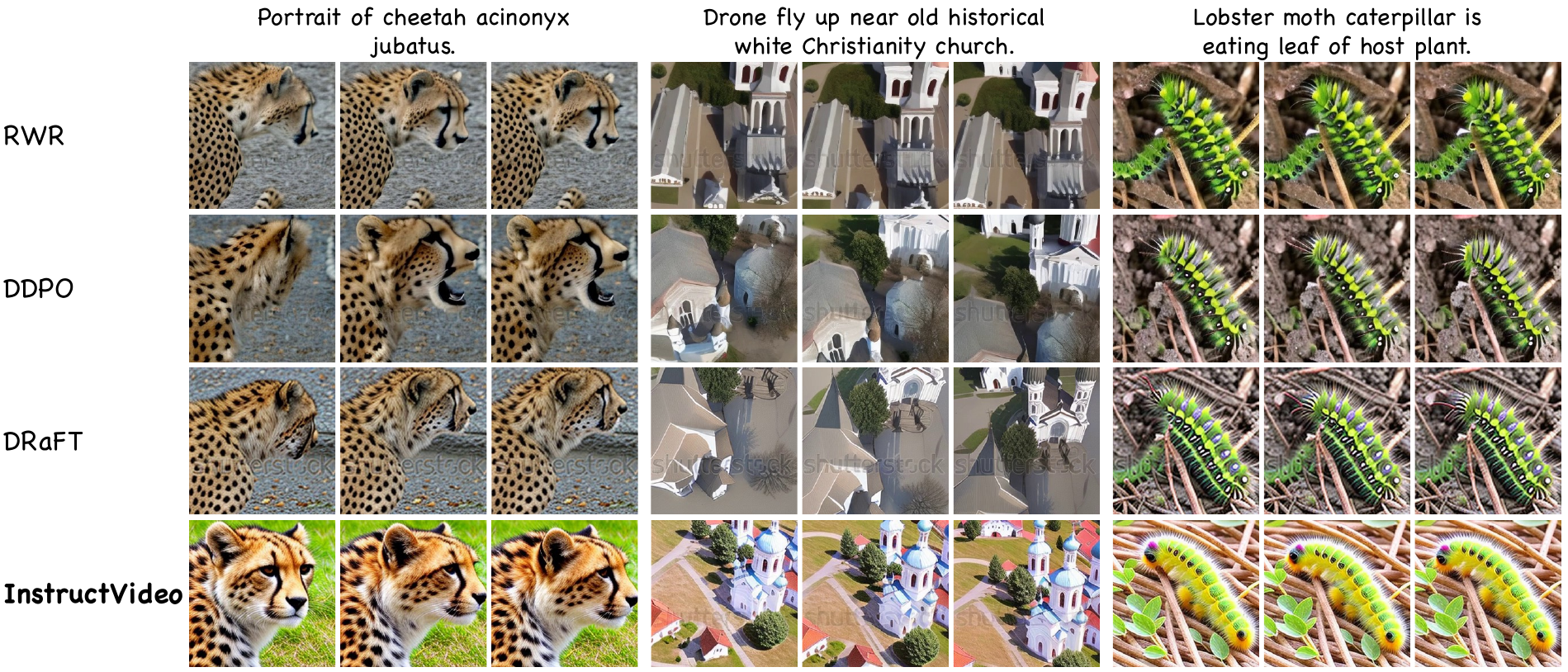}
    \vspace{-0.3cm}
    \caption{\small 
        More examples showing the \textbf{comparison of \method's generalization capabilities with other methods}.
        We set $D=20$ for all methods.
    }
\vspace{-0.4cm}
\label{fig:generalization-to-new-prompts-examples-supp}
\end{figure*}

We provide more visualization results to exemplify the conclusions we draw in the main paper, including:
\textbf{1)} More results demonstrating how the generated videos evolve as the fine-tuning process proceeds as shown in~\cref{fig:videos_evolve_supp};
\textbf{2)} More results showcasing the comparison between \method and the base model ModelScopeT2V as illustrated in~\cref{fig:comparison_with_ModelScopeT2V_supp};
\textbf{3)} More results exemplifying the comparison between \method and other reward fine-tuning methods as shown in~\cref{fig:comparing-with-other-reward-finetuning-examples-supp};
\textbf{4)} More results showing the \method's generalization capabilities to unseen text prompts as shown in~\cref{fig:generalization-to-new-prompts-examples-supp}.

\section{50-Step Generation with \method}\label{app:50-step_generation}
\begin{figure*}[t]
\centering
    \includegraphics[width=1.\textwidth]{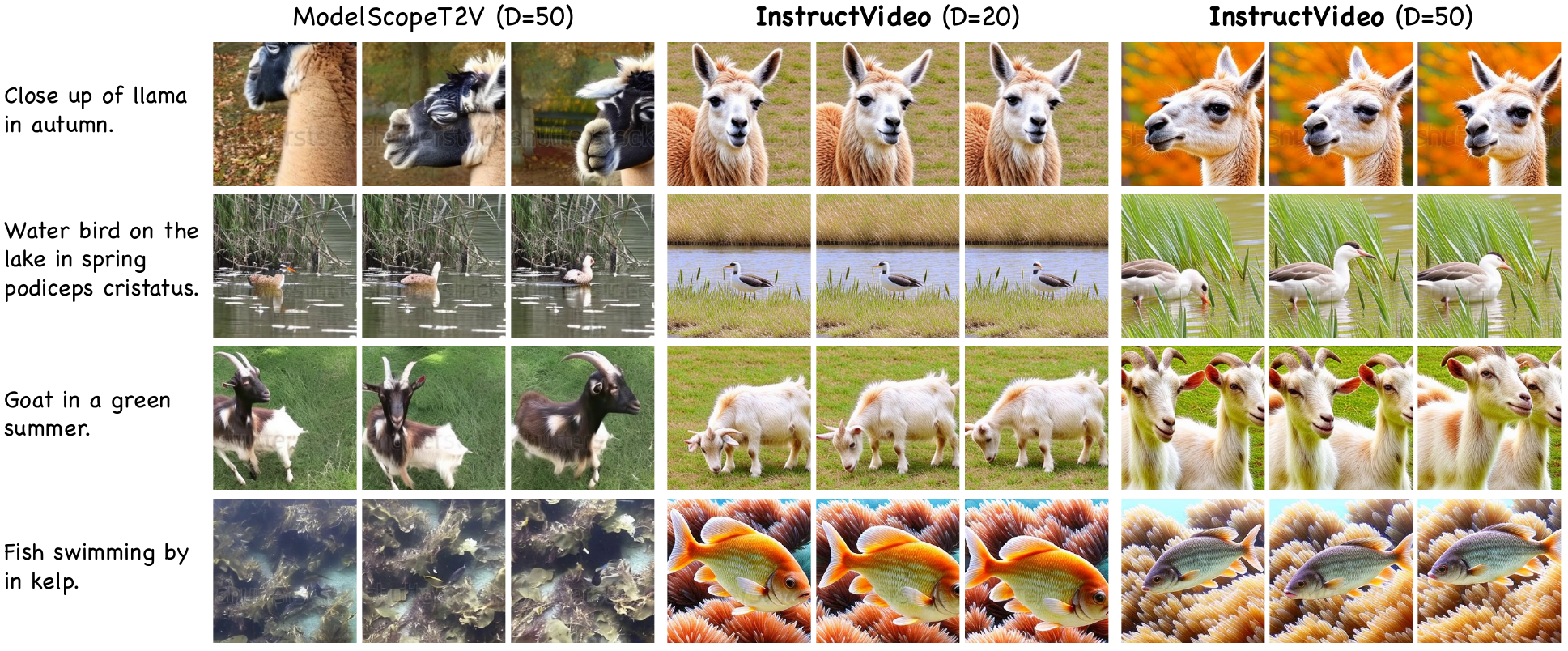}
    \vspace{-0.3cm}
    \caption{\small 
        \textbf{Generation with 50-step DDIM inference} after fine-tuning with 20-step DDIM inference.
    }
\vspace{-0.4cm}
\label{fig:50-step_ddim_inference-supp}
\end{figure*}

To showcase the adaptability and effectiveness of \method, we conduct experiments using a 50-step DDIM inference for generation after initial fine-tuning with 20-step DDIM inference.
The results are shown in~\cref{tab:generalization_to_new_prompts_supp}.
We observe that \method, despite being fine-tuned with a 20-step protocol, remains effective under a longer 50-step DDIM inference protocol, as demonstrated by the boosted reward scores.
We present several cases to further illustrate \method's efficacy as shown in~\cref{fig:50-step_ddim_inference-supp}.
We observe that both inference schemes can significantly improve over the base model and adopting more inference steps can occasionally lead to better results.

\begin{table}[t]
  \small
  \setlength{\tabcolsep}{2pt}
  \centering
    \begin{tabular}{c|ccc}
    \textbf{Method} & \textbf{In-domain} & \textbf{New Animals} & \textbf{Non-animals} \\
    \shline
    ModelScopeT2V & 0.2506 \tiny{± 0.0155} & 0.2502 \tiny{± 0.0138} & 0.2557 \tiny{± 0.0177} \\
    ModelScopeT2V\textsuperscript{\dag} & 0.2542 \tiny{± 0.0122} & 0.2541 \tiny{± 0.0109} & 0.2610 \tiny{± 0.0158} \\
    \hline
    \textbf{\method} & \textbf{0.2717 \tiny{± 0.0137}} & \textbf{0.2645 \tiny{± 0.0125}} & \textbf{0.2682 \tiny{± 0.0202}} \\
    \textbf{\method}\textsuperscript{\dag} & \textbf{0.2736 \tiny{± 0.0125}} & \textbf{0.2664 \tiny{± 0.0131}} & \textbf{0.2739 \tiny{± 0.0210}} \\
    \end{tabular}%
    \vspace{-0.2cm}
    \caption{
        \textbf{Generation with 50-step DDIM inference} after fine-tuning with 20-step DDIM inference.
        \dag \ denotes the model utilizes $D=50$ while others adopt $D=20$.
        `In-domain' denotes in-domain animal prompts from the evaluation data.
    }
    \vspace{-0.2cm}
  \label{tab:generalization_to_new_prompts_supp}%
\end{table}%

\section{50-Step Reward Fine-tuning}\label{app:50-step_reward_fine-tuning}
\begin{figure}[t]
\centering
    \includegraphics[width=.45\textwidth]{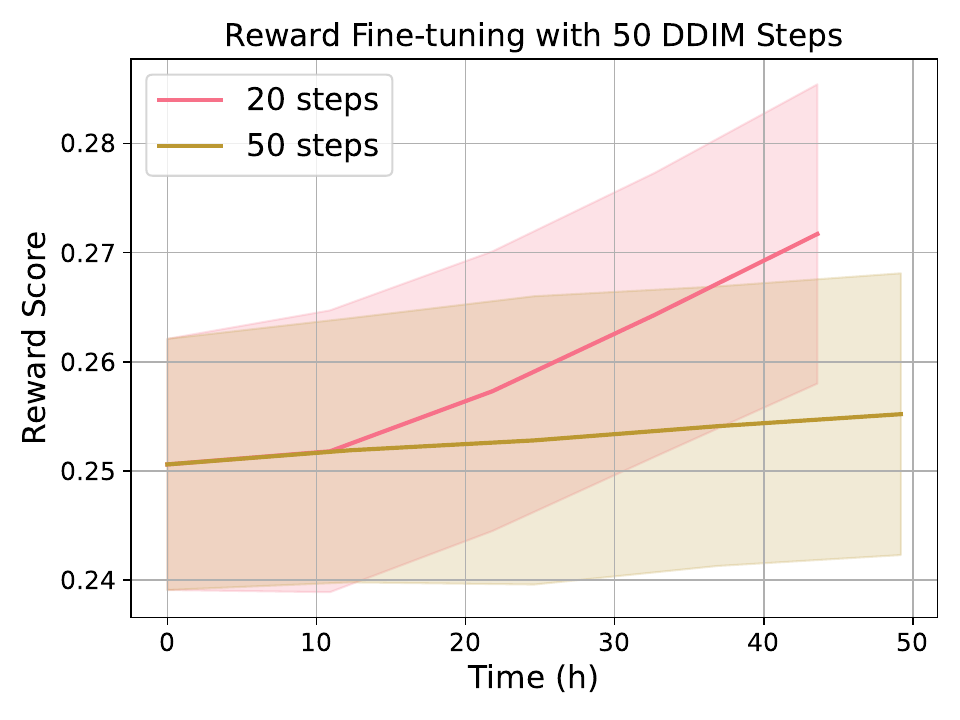}
    \vspace{-0.4cm}
    \caption{\small 
        \textbf{Reward finetuning with 50-step DDIM inference.}
    }
\vspace{-0.3cm}
\label{fig:reward_finetuning_with_50_ddim_steps}
\end{figure}

To assess the adaptation of \method to different DDIM steps, we experiment on reward fine-tuning with the commonly-used 50-step DDIM inference and evaluate its 20-step generation quality for a fair comparison.
We present the results in~\cref{fig:reward_finetuning_with_50_ddim_steps}.
The results demonstrate that \method could be optimized towards higher reward scores in both settings.
However, utilizing 50 steps degrades the fine-tuning efficiency, likely due to the increased computation brought by longer sampling chains. 

\section{Adaptation to Other Reward Functions}\label{app:adaptation_to_other_reward_functions}
\begin{figure*}[t]
\centering
    \includegraphics[width=1.\textwidth]{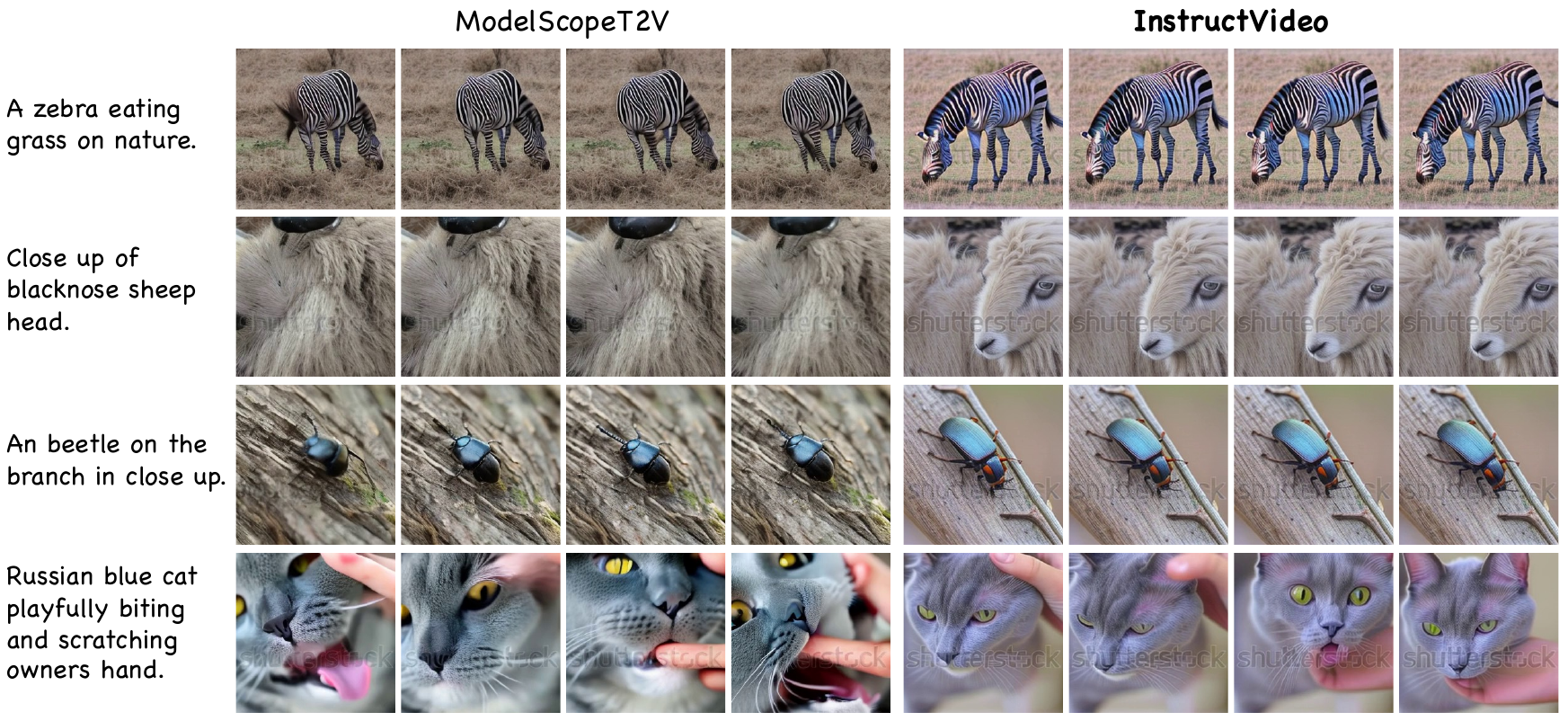}
    \vspace{-0.3cm}
    \caption{\small 
        \textbf{Comparison of \method fine-tuned using ImageReward with the base model ModelScopeT2V.
        We set $D=20$ for two methods.
        }
    }
\vspace{-0.4cm}
\label{fig:finetuning-with-ImageReward-supp}
\end{figure*}

In the main paper, we focus on the utilization of HPSv2~\cite{wu2023HPSv2}.
To further validate the generalization of our method to other reward functions, we explore the application of ImageReward~\cite{xu2023ImageReward} as our reward model.
ImageReward is a general-purpose text-to-image human preference
reward model, fine-tuned on BLIP~\cite{li2022blip}.
We perform reward fine-tuning as HPSv2 and present results in~\cref{fig:finetuning-with-ImageReward-supp}.
We observe that the quality of the videos are generally boosted in terms of structures, color vibrancy and details, despite that the stylistic aspects of the videos differ from those fine-tuned with HPSv2.


%% file: main-for-arXiv.bbl
\begin{thebibliography}{105}
\providecommand{\natexlab}[1]{#1}
\providecommand{\url}[1]{\texttt{#1}}
\expandafter\ifx\csname urlstyle\endcsname\relax
  \providecommand{\doi}[1]{doi: #1}\else
  \providecommand{\doi}{doi: \begingroup \urlstyle{rm}\Url}\fi

\bibitem[An et~al.(2023)An, Zhang, Yang, Gupta, Huang, Luo, and Yin]{an2023latent-shift}
Jie An, Songyang Zhang, Harry Yang, Sonal Gupta, Jia-Bin Huang, Jiebo Luo, and Xi Yin.
\newblock Latent-shift: Latent diffusion with temporal shift for efficient text-to-video generation.
\newblock \emph{arXiv preprint arXiv:2304.08477}, 2023.

\bibitem[Askell et~al.(2021)Askell, Bai, Chen, Drain, Ganguli, Henighan, Jones, Joseph, Mann, DasSarma, et~al.]{askell2021general_language_alignment}
Amanda Askell, Yuntao Bai, Anna Chen, Dawn Drain, Deep Ganguli, Tom Henighan, Andy Jones, Nicholas Joseph, Ben Mann, Nova DasSarma, et~al.
\newblock A general language assistant as a laboratory for alignment.
\newblock \emph{arXiv preprint arXiv:2112.00861}, 2021.

\bibitem[Bain et~al.(2021)Bain, Nagrani, Varol, and Zisserman]{Bain21Webvid}
Max Bain, Arsha Nagrani, G{\"u}l Varol, and Andrew Zisserman.
\newblock Frozen in time: A joint video and image encoder for end-to-end retrieval.
\newblock In \emph{IEEE International Conference on Computer Vision}, 2021.

\bibitem[Black et~al.(2023)Black, Janner, Du, Kostrikov, and Levine]{black2023DDPO}
Kevin Black, Michael Janner, Yilun Du, Ilya Kostrikov, and Sergey Levine.
\newblock Training diffusion models with reinforcement learning.
\newblock \emph{arXiv preprint arXiv:2305.13301}, 2023.

\bibitem[Blattmann et~al.(2023)Blattmann, Rombach, Ling, Dockhorn, Kim, Fidler, and Kreis]{blattmann2023align}
Andreas Blattmann, Robin Rombach, Huan Ling, Tim Dockhorn, Seung~Wook Kim, Sanja Fidler, and Karsten Kreis.
\newblock Align your latents: High-resolution video synthesis with latent diffusion models.
\newblock In \emph{Proceedings of the IEEE/CVF Conference on Computer Vision and Pattern Recognition}, pages 22563--22575, 2023.

\bibitem[Ceylan et~al.(2023)Ceylan, Huang, and Mitra]{ceylan2023Pix2video}
Duygu Ceylan, Chun-Hao~P Huang, and Niloy~J Mitra.
\newblock Pix2video: Video editing using image diffusion.
\newblock In \emph{Proceedings of the IEEE/CVF International Conference on Computer Vision}, pages 23206--23217, 2023.

\bibitem[Chai et~al.(2023)Chai, Guo, Wang, and Lu]{chai2023Stablevideo}
Wenhao Chai, Xun Guo, Gaoang Wang, and Yan Lu.
\newblock Stablevideo: Text-driven consistency-aware diffusion video editing.
\newblock In \emph{Proceedings of the IEEE/CVF International Conference on Computer Vision}, pages 23040--23050, 2023.

\bibitem[Chen et~al.(2023{\natexlab{a}})Chen, Wang, Zeng, Zhang, Zhou, Han, and Zhu]{chen2023videodreamer}
Hong Chen, Xin Wang, Guanning Zeng, Yipeng Zhang, Yuwei Zhou, Feilin Han, and Wenwu Zhu.
\newblock Videodreamer: Customized multi-subject text-to-video generation with disen-mix finetuning.
\newblock \emph{arXiv preprint arXiv:2311.00990}, 2023{\natexlab{a}}.

\bibitem[Chen et~al.(2023{\natexlab{b}})Chen, Xia, He, Zhang, Cun, Yang, Xing, Liu, Chen, Wang, Weng, and Shan]{chen2023videocrafter1}
Haoxin Chen, Menghan Xia, Yingqing He, Yong Zhang, Xiaodong Cun, Shaoshu Yang, Jinbo Xing, Yaofang Liu, Qifeng Chen, Xintao Wang, Chao Weng, and Ying Shan.
\newblock Videocrafter1: Open diffusion models for high-quality video generation, 2023{\natexlab{b}}.

\bibitem[Chen et~al.(2023{\natexlab{c}})Chen, Lin, Tseng, Lin, and Yang]{chen2023MCDiff}
Tsai-Shien Chen, Chieh~Hubert Lin, Hung-Yu Tseng, Tsung-Yi Lin, and Ming-Hsuan Yang.
\newblock Motion-conditioned diffusion model for controllable video synthesis.
\newblock \emph{arXiv preprint arXiv:2304.14404}, 2023{\natexlab{c}}.

\bibitem[Chen et~al.(2023{\natexlab{d}})Chen, Wu, Xie, Wu, Li, Xia, Xiao, and Lin]{chen2023controlavideo}
Weifeng Chen, Jie Wu, Pan Xie, Hefeng Wu, Jiashi Li, Xin Xia, Xuefeng Xiao, and Liang Lin.
\newblock Control-a-video: Controllable text-to-video generation with diffusion models, 2023{\natexlab{d}}.

\bibitem[Chen et~al.(2023{\natexlab{e}})Chen, Wang, Zhang, Zhuang, Ma, Yu, Wang, Lin, Qiao, and Liu]{chen2023seine}
Xinyuan Chen, Yaohui Wang, Lingjun Zhang, Shaobin Zhuang, Xin Ma, Jiashuo Yu, Yali Wang, Dahua Lin, Yu Qiao, and Ziwei Liu.
\newblock Seine: Short-to-long video diffusion model for generative transition and prediction.
\newblock \emph{arXiv preprint arXiv:2310.20700}, 2023{\natexlab{e}}.

\bibitem[Christiano et~al.(2017{\natexlab{a}})Christiano, Leike, Brown, Martic, Legg, and Amodei]{christiano2017deep}
Paul~F Christiano, Jan Leike, Tom Brown, Miljan Martic, Shane Legg, and Dario Amodei.
\newblock Deep reinforcement learning from human preferences.
\newblock \emph{Advances in neural information processing systems}, 30, 2017{\natexlab{a}}.

\bibitem[Christiano et~al.(2017{\natexlab{b}})Christiano, Leike, Brown, Martic, Legg, and Amodei]{christiano2017deep_RLHF}
Paul~F Christiano, Jan Leike, Tom Brown, Miljan Martic, Shane Legg, and Dario Amodei.
\newblock Deep reinforcement learning from human preferences.
\newblock \emph{Advances in neural information processing systems}, 30, 2017{\natexlab{b}}.

\bibitem[{\c{C}}i{\c{c}}ek et~al.(2016){\c{C}}i{\c{c}}ek, Abdulkadir, Lienkamp, Brox, and Ronneberger]{cciccek20163dUNet}
{\"O}zg{\"u}n {\c{C}}i{\c{c}}ek, Ahmed Abdulkadir, Soeren~S Lienkamp, Thomas Brox, and Olaf Ronneberger.
\newblock 3d u-net: learning dense volumetric segmentation from sparse annotation.
\newblock In \emph{Medical Image Computing and Computer-Assisted Intervention--MICCAI 2016: 19th International Conference, Athens, Greece, October 17-21, 2016, Proceedings, Part II 19}, pages 424--432. Springer, 2016.

\bibitem[Clark et~al.(2023)Clark, Vicol, Swersky, and Fleet]{clark2023DRaFT}
Kevin Clark, Paul Vicol, Kevin Swersky, and David~J Fleet.
\newblock Directly fine-tuning diffusion models on differentiable rewards.
\newblock \emph{arXiv preprint arXiv:2309.17400}, 2023.

\bibitem[Couairon et~al.(2022)Couairon, Verbeek, Schwenk, and Cord]{couairon2022Diffedit}
Guillaume Couairon, Jakob Verbeek, Holger Schwenk, and Matthieu Cord.
\newblock Diffedit: Diffusion-based semantic image editing with mask guidance.
\newblock \emph{arXiv preprint arXiv:2210.11427}, 2022.

\bibitem[Dong et~al.(2023)Dong, Xiong, Goyal, Pan, Diao, Zhang, Shum, and Zhang]{dong2023RAFT}
Hanze Dong, Wei Xiong, Deepanshu Goyal, Rui Pan, Shizhe Diao, Jipeng Zhang, Kashun Shum, and Tong Zhang.
\newblock Raft: Reward ranked finetuning for generative foundation model alignment.
\newblock \emph{arXiv preprint arXiv:2304.06767}, 2023.

\bibitem[Esser et~al.(2021)Esser, Rombach, and Ommer]{PatrickEsser2021TamingTF}
Patrick Esser, Robin Rombach, and Bj{\"o}rn Ommer.
\newblock Taming transformers for high-resolution image synthesis.
\newblock In \emph{Proceedings of the IEEE/CVF Conference on Computer Vision and Pattern Recognition}, 2021.

\bibitem[Esser et~al.(2023)Esser, Chiu, Atighehchian, Granskog, and Germanidis]{esser2023gen-1}
Patrick Esser, Johnathan Chiu, Parmida Atighehchian, Jonathan Granskog, and Anastasis Germanidis.
\newblock Structure and content-guided video synthesis with diffusion models.
\newblock \emph{arXiv preprint arXiv:2302.03011}, 2023.

\bibitem[Fan and Lee(2023)]{fan2023DDPM_shortcut_finetuning}
Ying Fan and Kangwook Lee.
\newblock Optimizing ddpm sampling with shortcut fine-tuning.
\newblock \emph{arXiv preprint arXiv:2301.13362}, 2023.

\bibitem[Fan et~al.(2023)Fan, Watkins, Du, Liu, Ryu, Boutilier, Abbeel, Ghavamzadeh, Lee, and Lee]{fan2023DPOK}
Ying Fan, Olivia Watkins, Yuqing Du, Hao Liu, Moonkyung Ryu, Craig Boutilier, Pieter Abbeel, Mohammad Ghavamzadeh, Kangwook Lee, and Kimin Lee.
\newblock Dpok: Reinforcement learning for fine-tuning text-to-image diffusion models.
\newblock \emph{arXiv preprint arXiv:2305.16381}, 2023.

\bibitem[Feng et~al.(2022{\natexlab{a}})Feng, Wang, and Yuan]{feng2022overcoming}
Tao Feng, Mang Wang, and Hangjie Yuan.
\newblock Overcoming catastrophic forgetting in incremental object detection via elastic response distillation.
\newblock In \emph{Proceedings of the IEEE/CVF Conference on Computer Vision and Pattern Recognition}, pages 9427--9436, 2022{\natexlab{a}}.

\bibitem[Feng et~al.(2022{\natexlab{b}})Feng, Yuan, Wang, Huang, Bian, and Zhang]{Feng2022PLwF}
Tao Feng, Hangjie Yuan, Mang Wang, Ziyuan Huang, Ang Bian, and Jianzhou Zhang.
\newblock Progressive learning without forgetting.
\newblock \emph{arXiv preprint arXiv:2211.15215}, 2022{\natexlab{b}}.

\bibitem[Fu et~al.(2023)Fu, Yu, Zhang, Fu, Su, Wang, and Bell]{fu2023tell_me}
Tsu-Jui Fu, Licheng Yu, Ning Zhang, Cheng-Yang Fu, Jong-Chyi Su, William~Yang Wang, and Sean Bell.
\newblock Tell me what happened: Unifying text-guided video completion via multimodal masked video generation.
\newblock In \emph{Proceedings of the IEEE/CVF Conference on Computer Vision and Pattern Recognition}, pages 10681--10692, 2023.

\bibitem[Ge et~al.(2023)Ge, Nah, Liu, Poon, Tao, Catanzaro, Jacobs, Huang, Liu, and Balaji]{ge2023preserve_correlation}
Songwei Ge, Seungjun Nah, Guilin Liu, Tyler Poon, Andrew Tao, Bryan Catanzaro, David Jacobs, Jia-Bin Huang, Ming-Yu Liu, and Yogesh Balaji.
\newblock Preserve your own correlation: A noise prior for video diffusion models.
\newblock In \emph{Proceedings of the IEEE/CVF International Conference on Computer Vision}, pages 22930--22941, 2023.

\bibitem[Geyer et~al.(2023)Geyer, Bar-Tal, Bagon, and Dekel]{geyer2023tokenflow}
Michal Geyer, Omer Bar-Tal, Shai Bagon, and Tali Dekel.
\newblock Tokenflow: Consistent diffusion features for consistent video editing.
\newblock \emph{arXiv preprint arXiv:2307.10373}, 2023.

\bibitem[Gruslys et~al.(2016)Gruslys, Munos, Danihelka, Lanctot, and Graves]{gruslys2016gradient_checkpointing}
Audrunas Gruslys, R{\'e}mi Munos, Ivo Danihelka, Marc Lanctot, and Alex Graves.
\newblock Memory-efficient backpropagation through time.
\newblock \emph{Advances in neural information processing systems}, 29, 2016.

\bibitem[Gu et~al.(2023)Gu, Wang, Zhao, Lu, Zhang, Wu, Xu, Zhang, Jiang, and Xu]{gu2023reuse_diffuse}
Jiaxi Gu, Shicong Wang, Haoyu Zhao, Tianyi Lu, Xing Zhang, Zuxuan Wu, Songcen Xu, Wei Zhang, Yu-Gang Jiang, and Hang Xu.
\newblock Reuse and diffuse: Iterative denoising for text-to-video generation.
\newblock \emph{arXiv preprint arXiv:2309.03549}, 2023.

\bibitem[He et~al.(2022)He, Yang, Zhang, Shan, and Chen]{he2022lvdm}
Yingqing He, Tianyu Yang, Yong Zhang, Ying Shan, and Qifeng Chen.
\newblock Latent video diffusion models for high-fidelity long video generation.
\newblock 2022.

\bibitem[He et~al.(2023)He, Xia, Chen, Cun, Gong, Xing, Zhang, Wang, Weng, Shan, et~al.]{he2023animate-a-story}
Yingqing He, Menghan Xia, Haoxin Chen, Xiaodong Cun, Yuan Gong, Jinbo Xing, Yong Zhang, Xintao Wang, Chao Weng, Ying Shan, et~al.
\newblock Animate-a-story: Storytelling with retrieval-augmented video generation.
\newblock \emph{arXiv preprint arXiv:2307.06940}, 2023.

\bibitem[Ho and Salimans(2022)]{ho2022classifier}
Jonathan Ho and Tim Salimans.
\newblock Classifier-free diffusion guidance.
\newblock \emph{arXiv preprint arXiv:2207.12598}, 2022.

\bibitem[Ho et~al.(2020)Ho, Jain, and Abbeel]{ho2020denoising_DDPM}
Jonathan Ho, Ajay Jain, and Pieter Abbeel.
\newblock Denoising diffusion probabilistic models.
\newblock \emph{Advances in Neural Information Processing Systems}, 33:\penalty0 6840--6851, 2020.

\bibitem[Ho et~al.(2022{\natexlab{a}})Ho, Chan, Saharia, Whang, Gao, Gritsenko, Kingma, Poole, Norouzi, Fleet, et~al.]{ho2022imagenvideo}
Jonathan Ho, William Chan, Chitwan Saharia, Jay Whang, Ruiqi Gao, Alexey Gritsenko, Diederik~P Kingma, Ben Poole, Mohammad Norouzi, David~J Fleet, et~al.
\newblock Imagen video: High definition video generation with diffusion models.
\newblock \emph{arXiv preprint arXiv:2210.02303}, 2022{\natexlab{a}}.

\bibitem[Ho et~al.(2022{\natexlab{b}})Ho, Salimans, Gritsenko, Chan, Norouzi, and Fleet]{ho2022video_diffusion_models}
Jonathan Ho, Tim Salimans, Alexey Gritsenko, William Chan, Mohammad Norouzi, and David~J Fleet.
\newblock Video diffusion models.
\newblock \emph{arXiv preprint arXiv:2204.03458}, 2022{\natexlab{b}}.

\bibitem[Hong et~al.(2022)Hong, Ding, Zheng, Liu, and Tang]{hong2022cogvideo}
Wenyi Hong, Ming Ding, Wendi Zheng, Xinghan Liu, and Jie Tang.
\newblock Cogvideo: Large-scale pretraining for text-to-video generation via transformers.
\newblock \emph{arXiv preprint arXiv:2205.15868}, 2022.

\bibitem[Hu et~al.(2021)Hu, Shen, Wallis, Allen-Zhu, Li, Wang, Wang, and Chen]{hu2021lora}
Edward~J Hu, Yelong Shen, Phillip Wallis, Zeyuan Allen-Zhu, Yuanzhi Li, Shean Wang, Lu Wang, and Weizhu Chen.
\newblock Lora: Low-rank adaptation of large language models.
\newblock \emph{arXiv preprint arXiv:2106.09685}, 2021.

\bibitem[Khachatryan et~al.(2023)Khachatryan, Movsisyan, Tadevosyan, Henschel, Wang, Navasardyan, and Shi]{khachatryan2023text2video-zero}
Levon Khachatryan, Andranik Movsisyan, Vahram Tadevosyan, Roberto Henschel, Zhangyang Wang, Shant Navasardyan, and Humphrey Shi.
\newblock Text2video-zero: Text-to-image diffusion models are zero-shot video generators.
\newblock \emph{arXiv preprint arXiv:2303.13439}, 2023.

\bibitem[Kirkpatrick et~al.(2017)Kirkpatrick, Pascanu, Rabinowitz, Veness, Desjardins, Rusu, Milan, Quan, Ramalho, Grabska-Barwinska, et~al.]{kirkpatrick2017overcomingforgetting}
James Kirkpatrick, Razvan Pascanu, Neil Rabinowitz, Joel Veness, Guillaume Desjardins, Andrei~A Rusu, Kieran Milan, John Quan, Tiago Ramalho, Agnieszka Grabska-Barwinska, et~al.
\newblock Overcoming catastrophic forgetting in neural networks.
\newblock \emph{Proceedings of the national academy of sciences}, 114\penalty0 (13):\penalty0 3521--3526, 2017.

\bibitem[Kirstain et~al.(2023)Kirstain, Polyak, Singer, Matiana, Penna, and Levy]{kirstain2023Pick-a-pic}
Yuval Kirstain, Adam Polyak, Uriel Singer, Shahbuland Matiana, Joe Penna, and Omer Levy.
\newblock Pick-a-pic: An open dataset of user preferences for text-to-image generation.
\newblock \emph{arXiv preprint arXiv:2305.01569}, 2023.

\bibitem[Kong et~al.(2021)Kong, Ping, Huang, Zhao, and Catanzaro]{kong2021diffwave}
Zhifeng Kong, Wei Ping, Jiaji Huang, Kexin Zhao, and Bryan Catanzaro.
\newblock Diffwave: A versatile diffusion model for audio synthesis.
\newblock In \emph{International Conference on Learning Representations}, 2021.

\bibitem[Lapid et~al.(2023)Lapid, Achituve, Bracha, and Fetaya]{lapid2023gd-vdm}
Ariel Lapid, Idan Achituve, Lior Bracha, and Ethan Fetaya.
\newblock Gd-vdm: Generated depth for better diffusion-based video generation.
\newblock \emph{arXiv preprint arXiv:2306.11173}, 2023.

\bibitem[Lee et~al.(2023)Lee, Liu, Ryu, Watkins, Du, Boutilier, Abbeel, Ghavamzadeh, and Gu]{lee2023AlignT2I}
Kimin Lee, Hao Liu, Moonkyung Ryu, Olivia Watkins, Yuqing Du, Craig Boutilier, Pieter Abbeel, Mohammad Ghavamzadeh, and Shixiang~Shane Gu.
\newblock Aligning text-to-image models using human feedback.
\newblock \emph{arXiv preprint arXiv:2302.12192}, 2023.

\bibitem[Leike et~al.(2018)Leike, Krueger, Everitt, Martic, Maini, and Legg]{leike2018scalable_agent_align}
Jan Leike, David Krueger, Tom Everitt, Miljan Martic, Vishal Maini, and Shane Legg.
\newblock Scalable agent alignment via reward modeling: a research direction.
\newblock \emph{arXiv preprint arXiv:1811.07871}, 2018.

\bibitem[Levine et~al.(2020)Levine, Kumar, Tucker, and Fu]{levine2020offlineRL}
Sergey Levine, Aviral Kumar, George Tucker, and Justin Fu.
\newblock Offline reinforcement learning: Tutorial, review, and perspectives on open problems.
\newblock \emph{arXiv preprint arXiv:2005.01643}, 2020.

\bibitem[Li et~al.(2023)Li, Zhang, Wu, Sun, Min, Liu, Zhai, and Lin]{li2023AGIQA-3K}
Chunyi Li, Zicheng Zhang, Haoning Wu, Wei Sun, Xiongkuo Min, Xiaohong Liu, Guangtao Zhai, and Weisi Lin.
\newblock Agiqa-3k: An open database for ai-generated image quality assessment.
\newblock \emph{arXiv preprint arXiv:2306.04717}, 2023.

\bibitem[Li et~al.(2022)Li, Li, Xiong, and Hoi]{li2022blip}
Junnan Li, Dongxu Li, Caiming Xiong, and Steven Hoi.
\newblock Blip: Bootstrapping language-image pre-training for unified vision-language understanding and generation.
\newblock In \emph{ICML}, pages 12888--12900. PMLR, 2022.

\bibitem[Li et~al.(2018)Li, Min, Shen, Carlson, and Carin]{li2018video_GAN_VAE}
Yitong Li, Martin Min, Dinghan Shen, David Carlson, and Lawrence Carin.
\newblock Video generation from text.
\newblock In \emph{Proceedings of the AAAI conference on artificial intelligence}, 2018.

\bibitem[Liu et~al.(2023{\natexlab{a}})Liu, Liu, Dai, Zeng, Cui, and Yang]{liu2023dual_T2V}
Binhui Liu, Xin Liu, Anbo Dai, Zhiyong Zeng, Zhen Cui, and Jian Yang.
\newblock Dual-stream diffusion net for text-to-video generation.
\newblock \emph{arXiv preprint arXiv:2308.08316}, 2023{\natexlab{a}}.

\bibitem[Liu et~al.(2023{\natexlab{b}})Liu, Zhang, Li, Lin, and Jia]{liu2023video-p2p}
Shaoteng Liu, Yuechen Zhang, Wenbo Li, Zhe Lin, and Jiaya Jia.
\newblock Video-p2p: Video editing with cross-attention control.
\newblock \emph{arXiv preprint arXiv:2303.04761}, 2023{\natexlab{b}}.

\bibitem[Ma et~al.(2023{\natexlab{a}})Ma, Xu, Li, Geng, Wang, and Wang]{ma2023optimal-noise-pursuit}
Shijie Ma, Huayi Xu, Mengjian Li, Weidong Geng, Meng Wang, and Yaxiong Wang.
\newblock Optimal noise pursuit for augmenting text-to-video generation.
\newblock \emph{arXiv preprint arXiv:2311.00949}, 2023{\natexlab{a}}.

\bibitem[Ma et~al.(2023{\natexlab{b}})Ma, He, Cun, Wang, Shan, Li, and Chen]{ma2023follow_your_pose}
Yue Ma, Yingqing He, Xiaodong Cun, Xintao Wang, Ying Shan, Xiu Li, and Qifeng Chen.
\newblock Follow your pose: Pose-guided text-to-video generation using pose-free videos.
\newblock \emph{arXiv preprint arXiv:2304.01186}, 2023{\natexlab{b}}.

\bibitem[Meng et~al.(2021)Meng, He, Song, Song, Wu, Zhu, and Ermon]{meng2021SDedit}
Chenlin Meng, Yutong He, Yang Song, Jiaming Song, Jiajun Wu, Jun-Yan Zhu, and Stefano Ermon.
\newblock Sdedit: Guided image synthesis and editing with stochastic differential equations.
\newblock \emph{arXiv preprint arXiv:2108.01073}, 2021.

\bibitem[Mittal et~al.(2017)Mittal, Marwah, and Balasubramanian]{mittal2017sync-draw}
Gaurav Mittal, Tanya Marwah, and Vineeth~N Balasubramanian.
\newblock Sync-draw: Automatic video generation using deep recurrent attentive architectures.
\newblock In \emph{Proceedings of the 25th ACM international conference on Multimedia}, pages 1096--1104, 2017.

\bibitem[Molad et~al.(2023)Molad, Horwitz, Valevski, Acha, Matias, Pritch, Leviathan, and Hoshen]{molad2023dreamix}
Eyal Molad, Eliahu Horwitz, Dani Valevski, Alex~Rav Acha, Yossi Matias, Yael Pritch, Yaniv Leviathan, and Yedid Hoshen.
\newblock Dreamix: Video diffusion models are general video editors.
\newblock \emph{arXiv preprint arXiv:2302.01329}, 2023.

\bibitem[OpenAI(2023)]{openai2023gpt4}
OpenAI.
\newblock {GPT-4} technical report, 2023.

\bibitem[Ouyang et~al.(2022)Ouyang, Wu, Jiang, Almeida, Wainwright, Mishkin, Zhang, Agarwal, Slama, Ray, et~al.]{ouyang2022InstructGPT}
Long Ouyang, Jeffrey Wu, Xu Jiang, Diogo Almeida, Carroll Wainwright, Pamela Mishkin, Chong Zhang, Sandhini Agarwal, Katarina Slama, Alex Ray, et~al.
\newblock Training language models to follow instructions with human feedback.
\newblock \emph{Advances in Neural Information Processing Systems}, 35:\penalty0 27730--27744, 2022.

\bibitem[Pan et~al.(2017)Pan, Qiu, Yao, Li, and Mei]{pan2017TGANs-C}
Yingwei Pan, Zhaofan Qiu, Ting Yao, Houqiang Li, and Tao Mei.
\newblock To create what you tell: Generating videos from captions.
\newblock In \emph{Proceedings of the 25th ACM international conference on Multimedia}, pages 1789--1798, 2017.

\bibitem[Prabhudesai et~al.(2023)Prabhudesai, Goyal, Pathak, and Fragkiadaki]{prabhudesai2023AlignProp}
Mihir Prabhudesai, Anirudh Goyal, Deepak Pathak, and Katerina Fragkiadaki.
\newblock Aligning text-to-image diffusion models with reward backpropagation.
\newblock \emph{arXiv preprint arXiv:2310.03739}, 2023.

\bibitem[Qi et~al.(2023)Qi, Cun, Zhang, Lei, Wang, Shan, and Chen]{qi2023fatezero}
Chenyang Qi, Xiaodong Cun, Yong Zhang, Chenyang Lei, Xintao Wang, Ying Shan, and Qifeng Chen.
\newblock Fatezero: Fusing attentions for zero-shot text-based video editing.
\newblock \emph{arXiv preprint arXiv:2303.09535}, 2023.

\bibitem[Qing et~al.(2023)Qing, Zhang, Wang, Wang, Wei, Zhang, Gao, and Sang]{qing2023HiGen}
Zhiwu Qing, Shiwei Zhang, Jiayu Wang, Xiang Wang, Yujie Wei, Yingya Zhang, Changxin Gao, and Nong Sang.
\newblock Hierarchical spatio-temporal decoupling for text-to-video generation.
\newblock \emph{arXiv preprint arXiv:2312.04483}, 2023.

\bibitem[Radford et~al.(2021)Radford, Kim, Hallacy, Ramesh, Goh, Agarwal, Sastry, Askell, Mishkin, Clark, et~al.]{radford2021CLIP}
Alec Radford, Jong~Wook Kim, Chris Hallacy, Aditya Ramesh, Gabriel Goh, Sandhini Agarwal, Girish Sastry, Amanda Askell, Pamela Mishkin, Jack Clark, et~al.
\newblock Learning transferable visual models from natural language supervision.
\newblock In \emph{International conference on machine learning}, pages 8748--8763. PMLR, 2021.

\bibitem[Rombach et~al.(2022)Rombach, Blattmann, Lorenz, Esser, and Ommer]{rombach2022stable_diffusion}
Robin Rombach, Andreas Blattmann, Dominik Lorenz, Patrick Esser, and Bj{\"o}rn Ommer.
\newblock High-resolution image synthesis with latent diffusion models.
\newblock In \emph{Proceedings of the IEEE/CVF Conference on Computer Vision and Pattern Recognition}, pages 10684--10695, 2022.

\bibitem[Ronneberger et~al.(2015)Ronneberger, Fischer, and Brox]{ronneberger2015UNet}
Olaf Ronneberger, Philipp Fischer, and Thomas Brox.
\newblock U-net: Convolutional networks for biomedical image segmentation.
\newblock In \emph{Medical Image Computing and Computer-Assisted Intervention--MICCAI 2015: 18th International Conference, Munich, Germany, October 5-9, 2015, Proceedings, Part III 18}, pages 234--241. Springer, 2015.

\bibitem[Ruan et~al.(2023)Ruan, Ma, Yang, He, Liu, Fu, Yuan, Jin, and Guo]{ruan2023mm-diffusion}
Ludan Ruan, Yiyang Ma, Huan Yang, Huiguo He, Bei Liu, Jianlong Fu, Nicholas~Jing Yuan, Qin Jin, and Baining Guo.
\newblock Mm-diffusion: Learning multi-modal diffusion models for joint audio and video generation.
\newblock In \emph{Proceedings of the IEEE/CVF Conference on Computer Vision and Pattern Recognition}, pages 10219--10228, 2023.

\bibitem[Schuhmann et~al.(2021)Schuhmann, Vencu, Beaumont, Kaczmarczyk, Mullis, Katta, Coombes, Jitsev, and Komatsuzaki]{schuhmann2021laion}
Christoph Schuhmann, Richard Vencu, Romain Beaumont, Robert Kaczmarczyk, Clayton Mullis, Aarush Katta, Theo Coombes, Jenia Jitsev, and Aran Komatsuzaki.
\newblock Laion-400m: Open dataset of clip-filtered 400 million image-text pairs.
\newblock \emph{arXiv preprint arXiv:2111.02114}, 2021.

\bibitem[Singer et~al.(2022)Singer, Polyak, Hayes, Yin, An, Zhang, Hu, Yang, Ashual, Gafni, et~al.]{singer2022make-a-video}
Uriel Singer, Adam Polyak, Thomas Hayes, Xi Yin, Jie An, Songyang Zhang, Qiyuan Hu, Harry Yang, Oron Ashual, Oran Gafni, et~al.
\newblock Make-a-video: Text-to-video generation without text-video data.
\newblock \emph{arXiv preprint arXiv:2209.14792}, 2022.

\bibitem[Skorokhodov et~al.(2022)Skorokhodov, Tulyakov, and Elhoseiny]{skorokhodov2022stylegan-v}
Ivan Skorokhodov, Sergey Tulyakov, and Mohamed Elhoseiny.
\newblock Stylegan-v: A continuous video generator with the price, image quality and perks of stylegan2.
\newblock In \emph{Proceedings of the IEEE/CVF Conference on Computer Vision and Pattern Recognition}, pages 3626--3636, 2022.

\bibitem[Sohl-Dickstein et~al.(2015)Sohl-Dickstein, Weiss, Maheswaranathan, and Ganguli]{sohl2015Diffusion_model}
Jascha Sohl-Dickstein, Eric Weiss, Niru Maheswaranathan, and Surya Ganguli.
\newblock Deep unsupervised learning using nonequilibrium thermodynamics.
\newblock In \emph{International Conference on Computer Vision}, pages 2256--2265. PMLR, 2015.

\bibitem[Song et~al.(2020{\natexlab{a}})Song, Meng, and Ermon]{song2020DDIM}
Jiaming Song, Chenlin Meng, and Stefano Ermon.
\newblock Denoising diffusion implicit models.
\newblock \emph{arXiv preprint arXiv:2010.02502}, 2020{\natexlab{a}}.

\bibitem[Song et~al.(2020{\natexlab{b}})Song, Sohl-Dickstein, Kingma, Kumar, Ermon, and Poole]{song2020score_generative_SDE}
Yang Song, Jascha Sohl-Dickstein, Diederik~P Kingma, Abhishek Kumar, Stefano Ermon, and Ben Poole.
\newblock Score-based generative modeling through stochastic differential equations.
\newblock \emph{arXiv preprint arXiv:2011.13456}, 2020{\natexlab{b}}.

\bibitem[Stiennon et~al.(2020)Stiennon, Ouyang, Wu, Ziegler, Lowe, Voss, Radford, Amodei, and Christiano]{stiennon2020summerize_HF}
Nisan Stiennon, Long Ouyang, Jeffrey Wu, Daniel Ziegler, Ryan Lowe, Chelsea Voss, Alec Radford, Dario Amodei, and Paul~F Christiano.
\newblock Learning to summarize with human feedback.
\newblock \emph{Advances in Neural Information Processing Systems}, 33:\penalty0 3008--3021, 2020.

\bibitem[Tang et~al.(2023)Tang, Yang, Zhu, Zeng, and Bansal]{tang2023any-to-any}
Zineng Tang, Ziyi Yang, Chenguang Zhu, Michael Zeng, and Mohit Bansal.
\newblock Any-to-any generation via composable diffusion.
\newblock \emph{arXiv preprint arXiv:2305.11846}, 2023.

\bibitem[Touvron et~al.(2023)Touvron, Lavril, Izacard, Martinet, Lachaux, Lacroix, Rozière, Goyal, Hambro, Azhar, Rodriguez, Joulin, Grave, and Lample]{touvron2023llama}
Hugo Touvron, Thibaut Lavril, Gautier Izacard, Xavier Martinet, Marie-Anne Lachaux, Timothée Lacroix, Baptiste Rozière, Naman Goyal, Eric Hambro, Faisal Azhar, Aurelien Rodriguez, Armand Joulin, Edouard Grave, and Guillaume Lample.
\newblock Llama: Open and efficient foundation language models, 2023.

\bibitem[Vaswani et~al.(2017)Vaswani, Shazeer, Parmar, Uszkoreit, Jones, Gomez, Kaiser, and Polosukhin]{AttentionAlluNeed}
Ashish Vaswani, Noam Shazeer, Niki Parmar, Jakob Uszkoreit, Llion Jones, Aidan~N Gomez, {\L}ukasz Kaiser, and Illia Polosukhin.
\newblock Attention is all you need.
\newblock In \emph{NeurIPS}, pages 5998--6008, 2017.

\bibitem[Villegas et~al.(2022)Villegas, Babaeizadeh, Kindermans, Moraldo, Zhang, Saffar, Castro, Kunze, and Erhan]{villegas2022phenaki_GAN}
Ruben Villegas, Mohammad Babaeizadeh, Pieter-Jan Kindermans, Hernan Moraldo, Han Zhang, Mohammad~Taghi Saffar, Santiago Castro, Julius Kunze, and Dumitru Erhan.
\newblock Phenaki: Variable length video generation from open domain textual description.
\newblock \emph{arXiv preprint arXiv:2210.02399}, 2022.

\bibitem[von R{\"u}tte et~al.(2023)von R{\"u}tte, Fedele, Thomm, and Wolf]{von2023FABRIC}
Dimitri von R{\"u}tte, Elisabetta Fedele, Jonathan Thomm, and Lukas Wolf.
\newblock Fabric: Personalizing diffusion models with iterative feedback.
\newblock \emph{arXiv preprint arXiv:2307.10159}, 2023.

\bibitem[Wang et~al.(2023{\natexlab{a}})Wang, Yuan, Chen, Zhang, Wang, and Zhang]{wang2023modelscopeT2V}
Jiuniu Wang, Hangjie Yuan, Dayou Chen, Yingya Zhang, Xiang Wang, and Shiwei Zhang.
\newblock Modelscope text-to-video technical report.
\newblock \emph{arXiv preprint arXiv:2308.06571}, 2023{\natexlab{a}}.

\bibitem[Wang et~al.(2016)Wang, Xiong, Wang, Qiao, Lin, Tang, and Van~Gool]{wang2016TSN}
Limin Wang, Yuanjun Xiong, Zhe Wang, Yu Qiao, Dahua Lin, Xiaoou Tang, and Luc Van~Gool.
\newblock Temporal segment networks: Towards good practices for deep action recognition.
\newblock In \emph{ECCV}, pages 20--36. Springer, 2016.

\bibitem[Wang et~al.(2023{\natexlab{b}})Wang, Yang, Tuo, He, Zhu, Fu, and Liu]{wang2023videofactory}
Wenjing Wang, Huan Yang, Zixi Tuo, Huiguo He, Junchen Zhu, Jianlong Fu, and Jiaying Liu.
\newblock Videofactory: Swap attention in spatiotemporal diffusions for text-to-video generation.
\newblock \emph{arXiv preprint arXiv:2305.10874}, 2023{\natexlab{b}}.

\bibitem[Wang et~al.(2023{\natexlab{c}})Wang, Yuan, Zhang, Chen, Wang, Zhang, Shen, Zhao, and Zhou]{2023videocomposer}
Xiang Wang, Hangjie Yuan, Shiwei Zhang, Dayou Chen, Jiuniu Wang, Yingya Zhang, Yujun Shen, Deli Zhao, and Jingren Zhou.
\newblock Videocomposer: Compositional video synthesis with motion controllability.
\newblock \emph{arXiv preprint arXiv:2306.02018}, 2023{\natexlab{c}}.

\bibitem[Wang et~al.(2023{\natexlab{d}})Wang, Zhang, Zhang, Liu, Zhang, Gao, and Sang]{wang2023videolcm}
Xiang Wang, Shiwei Zhang, Han Zhang, Yu Liu, Yingya Zhang, Changxin Gao, and Nong Sang.
\newblock Videolcm: Video latent consistency model.
\newblock \emph{arXiv preprint arXiv:2312.09109}, 2023{\natexlab{d}}.

\bibitem[Wang et~al.(2023{\natexlab{e}})Wang, Chen, Ma, Zhou, Huang, Wang, Yang, He, Yu, Yang, et~al.]{wang2023lavie}
Yaohui Wang, Xinyuan Chen, Xin Ma, Shangchen Zhou, Ziqi Huang, Yi Wang, Ceyuan Yang, Yinan He, Jiashuo Yu, Peiqing Yang, et~al.
\newblock Lavie: High-quality video generation with cascaded latent diffusion models.
\newblock \emph{arXiv preprint arXiv:2309.15103}, 2023{\natexlab{e}}.

\bibitem[Wei et~al.(2023)Wei, Zhang, Qing, Yuan, Liu, Liu, Zhang, Zhou, and Shan]{wei2023dreamvideo}
Yujie Wei, Shiwei Zhang, Zhiwu Qing, Hangjie Yuan, Zhiheng Liu, Yu Liu, Yingya Zhang, Jingren Zhou, and Hongming Shan.
\newblock Dreamvideo: Composing your dream videos with customized subject and motion.
\newblock \emph{arXiv preprint arXiv:2312.04433}, 2023.

\bibitem[Wu et~al.(2021)Wu, Huang, Zhang, Li, Ji, Yang, Sapiro, and Duan]{wu2021godiva}
Chenfei Wu, Lun Huang, Qianxi Zhang, Binyang Li, Lei Ji, Fan Yang, Guillermo Sapiro, and Nan Duan.
\newblock Godiva: Generating open-domain videos from natural descriptions.
\newblock \emph{arXiv preprint arXiv:2104.14806}, 2021.

\bibitem[Wu et~al.(2022)Wu, Liang, Ji, Yang, Fang, Jiang, and Duan]{wu2022nuwa}
Chenfei Wu, Jian Liang, Lei Ji, Fan Yang, Yuejian Fang, Daxin Jiang, and Nan Duan.
\newblock N{\"u}wa: Visual synthesis pre-training for neural visual world creation.
\newblock In \emph{European Conference on Computer Vision}, pages 720--736. Springer, 2022.

\bibitem[Wu et~al.(2023{\natexlab{a}})Wu, Ge, Wang, Lei, Gu, Shi, Hsu, Shan, Qie, and Shou]{wu2023tune-a-video}
Jay~Zhangjie Wu, Yixiao Ge, Xintao Wang, Stan~Weixian Lei, Yuchao Gu, Yufei Shi, Wynne Hsu, Ying Shan, Xiaohu Qie, and Mike~Zheng Shou.
\newblock Tune-a-video: One-shot tuning of image diffusion models for text-to-video generation.
\newblock In \emph{Proceedings of the IEEE/CVF International Conference on Computer Vision}, pages 7623--7633, 2023{\natexlab{a}}.

\bibitem[Wu et~al.(2023{\natexlab{b}})Wu, Hao, Sun, Chen, Zhu, Zhao, and Li]{wu2023HPSv2}
Xiaoshi Wu, Yiming Hao, Keqiang Sun, Yixiong Chen, Feng Zhu, Rui Zhao, and Hongsheng Li.
\newblock Human preference score v2: A solid benchmark for evaluating human preferences of text-to-image synthesis.
\newblock \emph{arXiv preprint arXiv:2306.09341}, 2023{\natexlab{b}}.

\bibitem[Wu et~al.(2023{\natexlab{c}})Wu, Sun, Zhu, Zhao, and Li]{wu2023HPSv1}
Xiaoshi Wu, Keqiang Sun, Feng Zhu, Rui Zhao, and Hongsheng Li.
\newblock Human preference score: Better aligning text-to-image models with human preference.
\newblock In \emph{Proceedings of the IEEE/CVF International Conference on Computer Vision}, pages 2096--2105, 2023{\natexlab{c}}.

\bibitem[Xing et~al.(2023{\natexlab{a}})Xing, Xia, Liu, Zhang, Zhang, He, Liu, Chen, Cun, Wang, et~al.]{xing2023make-your-video}
Jinbo Xing, Menghan Xia, Yuxin Liu, Yuechen Zhang, Yong Zhang, Yingqing He, Hanyuan Liu, Haoxin Chen, Xiaodong Cun, Xintao Wang, et~al.
\newblock Make-your-video: Customized video generation using textual and structural guidance.
\newblock \emph{arXiv preprint arXiv:2306.00943}, 2023{\natexlab{a}}.

\bibitem[Xing et~al.(2023{\natexlab{b}})Xing, Xia, Zhang, Chen, Wang, Wong, and Shan]{xing2023dynamicrafter}
Jinbo Xing, Menghan Xia, Yong Zhang, Haoxin Chen, Xintao Wang, Tien-Tsin Wong, and Ying Shan.
\newblock Dynamicrafter: Animating open-domain images with video diffusion priors.
\newblock 2023{\natexlab{b}}.

\bibitem[Xing et~al.(2023{\natexlab{c}})Xing, Dai, Hu, Wu, and Jiang]{xing2023SimDA}
Zhen Xing, Qi Dai, Han Hu, Zuxuan Wu, and Yu-Gang Jiang.
\newblock Simda: Simple diffusion adapter for efficient video generation.
\newblock \emph{arXiv preprint arXiv:2308.09710}, 2023{\natexlab{c}}.

\bibitem[Xu et~al.(2023)Xu, Liu, Wu, Tong, Li, Ding, Tang, and Dong]{xu2023ImageReward}
Jiazheng Xu, Xiao Liu, Yuchen Wu, Yuxuan Tong, Qinkai Li, Ming Ding, Jie Tang, and Yuxiao Dong.
\newblock Imagereward: Learning and evaluating human preferences for text-to-image generation.
\newblock \emph{arXiv preprint arXiv:2304.05977}, 2023.

\bibitem[Yan et~al.(2021)Yan, Zhang, Abbeel, and Srinivas]{yan2021videogpt}
Wilson Yan, Yunzhi Zhang, Pieter Abbeel, and Aravind Srinivas.
\newblock Videogpt: Video generation using vq-vae and transformers.
\newblock \emph{arXiv preprint arXiv:2104.10157}, 2021.

\bibitem[Yin et~al.(2023)Yin, Wu, Liang, Shi, Li, Ming, and Duan]{yin2023dragnuwa}
Shengming Yin, Chenfei Wu, Jian Liang, Jie Shi, Houqiang Li, Gong Ming, and Nan Duan.
\newblock Dragnuwa: Fine-grained control in video generation by integrating text, image, and trajectory.
\newblock \emph{arXiv preprint arXiv:2308.08089}, 2023.

\bibitem[Yu et~al.(2022)Yu, Tack, Mo, Kim, Kim, Ha, and Shin]{yu2022video_implicit_GAN}
Sihyun Yu, Jihoon Tack, Sangwoo Mo, Hyunsu Kim, Junho Kim, Jung-Woo Ha, and Jinwoo Shin.
\newblock Generating videos with dynamics-aware implicit generative adversarial networks.
\newblock \emph{arXiv preprint arXiv:2202.10571}, 2022.

\bibitem[Yu et~al.(2023)Yu, Sohn, Kim, and Shin]{yu2023PVDM}
Sihyun Yu, Kihyuk Sohn, Subin Kim, and Jinwoo Shin.
\newblock Video probabilistic diffusion models in projected latent space.
\newblock In \emph{Proceedings of the IEEE/CVF Conference on Computer Vision and Pattern Recognition}, pages 18456--18466, 2023.

\bibitem[Zhang et~al.(2023{\natexlab{a}})Zhang, Wu, Liu, Zhao, Ran, Gu, Gao, and Shou]{zhang2023show-1}
David~Junhao Zhang, Jay~Zhangjie Wu, Jia-Wei Liu, Rui Zhao, Lingmin Ran, Yuchao Gu, Difei Gao, and Mike~Zheng Shou.
\newblock Show-1: Marrying pixel and latent diffusion models for text-to-video generation, 2023{\natexlab{a}}.

\bibitem[Zhang et~al.(2022)Zhang, Tao, and Chen]{zhang2022gddim}
Qinsheng Zhang, Molei Tao, and Yongxin Chen.
\newblock gddim: Generalized denoising diffusion implicit models.
\newblock \emph{arXiv preprint arXiv:2206.05564}, 2022.

\bibitem[Zhang et~al.(2023{\natexlab{b}})Zhang, Wang, Zhang, Zhao, Yuan, Qin, Wang, Zhao, and Zhou]{zhang2023i2vgen-xl}
Shiwei Zhang, Jiayu Wang, Yingya Zhang, Kang Zhao, Hangjie Yuan, Zhiwu Qin, Xiang Wang, Deli Zhao, and Jingren Zhou.
\newblock I2vgen-xl: High-quality image-to-video synthesis via cascaded diffusion models.
\newblock \emph{arXiv preprint arXiv:2311.04145}, 2023{\natexlab{b}}.

\bibitem[Zhang et~al.(2023{\natexlab{c}})Zhang, Yang, Feng, Qin, Chen, Yu, Chen, Wang, Savarese, Ermon, et~al.]{zhang2023HIVE_feedback_edit}
Shu Zhang, Xinyi Yang, Yihao Feng, Can Qin, Chia-Chih Chen, Ning Yu, Zeyuan Chen, Huan Wang, Silvio Savarese, Stefano Ermon, et~al.
\newblock Hive: Harnessing human feedback for instructional visual editing.
\newblock \emph{arXiv preprint arXiv:2303.09618}, 2023{\natexlab{c}}.

\bibitem[Zhao et~al.(2023{\natexlab{a}})Zhao, Wang, Bao, Li, and Zhu]{zhao2023controlvideo}
Min Zhao, Rongzhen Wang, Fan Bao, Chongxuan Li, and Jun Zhu.
\newblock Controlvideo: Adding conditional control for one shot text-to-video editing.
\newblock \emph{arXiv preprint arXiv:2305.17098}, 2023{\natexlab{a}}.

\bibitem[Zhao et~al.(2023{\natexlab{b}})Zhao, Gu, Wu, Zhang, Liu, Wu, Keppo, and Shou]{zhao2023motiondirector}
Rui Zhao, Yuchao Gu, Jay~Zhangjie Wu, David~Junhao Zhang, Jiawei Liu, Weijia Wu, Jussi Keppo, and Mike~Zheng Shou.
\newblock Motiondirector: Motion customization of text-to-video diffusion models.
\newblock \emph{arXiv preprint arXiv:2310.08465}, 2023{\natexlab{b}}.

\bibitem[Zhou et~al.(2022)Zhou, Wang, Yan, Lv, Zhu, and Feng]{zhou2022magicvideo}
Daquan Zhou, Weimin Wang, Hanshu Yan, Weiwei Lv, Yizhe Zhu, and Jiashi Feng.
\newblock Magicvideo: Efficient video generation with latent diffusion models.
\newblock \emph{arXiv preprint arXiv:2211.11018}, 2022.

\bibitem[Zhu et~al.(2023)Zhu, Yang, He, Wang, Tuo, Cheng, Gao, Song, and Fu]{zhu2023moviefactory}
Junchen Zhu, Huan Yang, Huiguo He, Wenjing Wang, Zixi Tuo, Wen-Huang Cheng, Lianli Gao, Jingkuan Song, and Jianlong Fu.
\newblock Moviefactory: Automatic movie creation from text using large generative models for language and images.
\newblock \emph{arXiv preprint arXiv:2306.07257}, 2023.

\end{thebibliography}
